\documentclass{article}

\usepackage{microtype}
\usepackage{graphicx}
\usepackage{subcaption}
\usepackage{booktabs} 
\usepackage{multirow, makecell, colortbl, hhline}
\usepackage{arydshln} 
\usepackage{soul}
\usepackage[most]{tcolorbox}
\usepackage{siunitx} 
\usepackage{hyperref}

\usepackage[preprint]{icml2026}

\usepackage{amsmath}
\usepackage{amssymb}
\usepackage{mathtools}
\usepackage{amsthm}
\usepackage{array}
\usepackage{xurl}

\usepackage[capitalize,noabbrev]{cleveref}

\theoremstyle{plain}

\theoremstyle{definition}

\theoremstyle{remark}

\newcommand{\shamit}[1]{{#1}}

\newcommand{\modelname}[1]{{UNITE}}

\begin{document}

\twocolumn[






    \icmltitle{
    End-to-End Training for Unified Tokenization and Latent Denoising
    }







  \icmlsetsymbol{equal}{*}

  \begin{icmlauthorlist}
    \icmlauthor{Shivam Duggal}{xxx,equal} \quad
    \icmlauthor{Xingjian Bai}{xxx,equal} \quad
    \icmlauthor{Zongze Wu}{yyy} \quad
    \icmlauthor{Richard Zhang}{yyy} \quad
    \icmlauthor{Eli Shechtman}{yyy} \\
    \icmlauthor{Antonio Torralba}{xxx} \quad
    \icmlauthor{Phillip Isola}{xxx} \quad
    \icmlauthor{William T. Freeman}{xxx} \\
  \end{icmlauthorlist}

  \icmlaffiliation{xxx}{Massachusetts Institute of Technology}
  \icmlaffiliation{yyy}{Adobe}


  \icmlkeywords{Machine Learning, ICML}

  \vskip 0.2in
]




\printAffiliationsAndNotice{\icmlEqualContribution}

\begin{abstract}
    Latent diffusion models (LDMs) enable high-fidelity synthesis by operating in learned latent spaces.
    However, training state-of-the-art LDMs requires complex staging: a tokenizer must be trained first,
    before the diffusion model can be trained in the frozen latent space.
    We propose \modelname{} -- an autoencoder architecture for unified tokenization and latent diffusion.
    \modelname{} consists of a \emph{Generative Encoder} that serves as both image tokenizer and latent generator via weight sharing.
    Our key insight is that tokenization and generation can be viewed as the same latent inference problem under different conditioning regimes: tokenization infers latents from fully observed images, whereas generation infers them from noise together with text or class conditioning.
    Motivated by this, we introduce a single-stage training procedure that jointly optimizes both tasks via two forward passes through the same Generative Encoder.
    The shared parameters enable gradients to jointly shape the latent space, encouraging a ``common latent language''.
    Across image and molecule modalities, \modelname{} achieves near state-of-the-art performance without adversarial losses or any pretrained encoders (e.g., DINO), reaching FID 2.12 and 1.73 for Base and Large models on ImageNet $256 \times 256$. We further analyze the Generative Encoder through the lenses of representation alignment and compression.
    These results show that single-stage joint training of tokenization and generation from scratch is feasible. \\ \\
    {\textbf{Code:} \hypersetup{urlcolor=magenta}\url{https://github.com/ShivamDuggal4/UNITE-tokenization-generation}} \\
    {\textbf{Project Page:} \hypersetup{urlcolor=magenta}\url{https://xingjianbai.com/unite-tokenization-generation/}}

    \vspace{-3mm}
\end{abstract}

\section{Introduction}

\begin{figure}[t]
    \centering
    \includegraphics[width=\columnwidth]{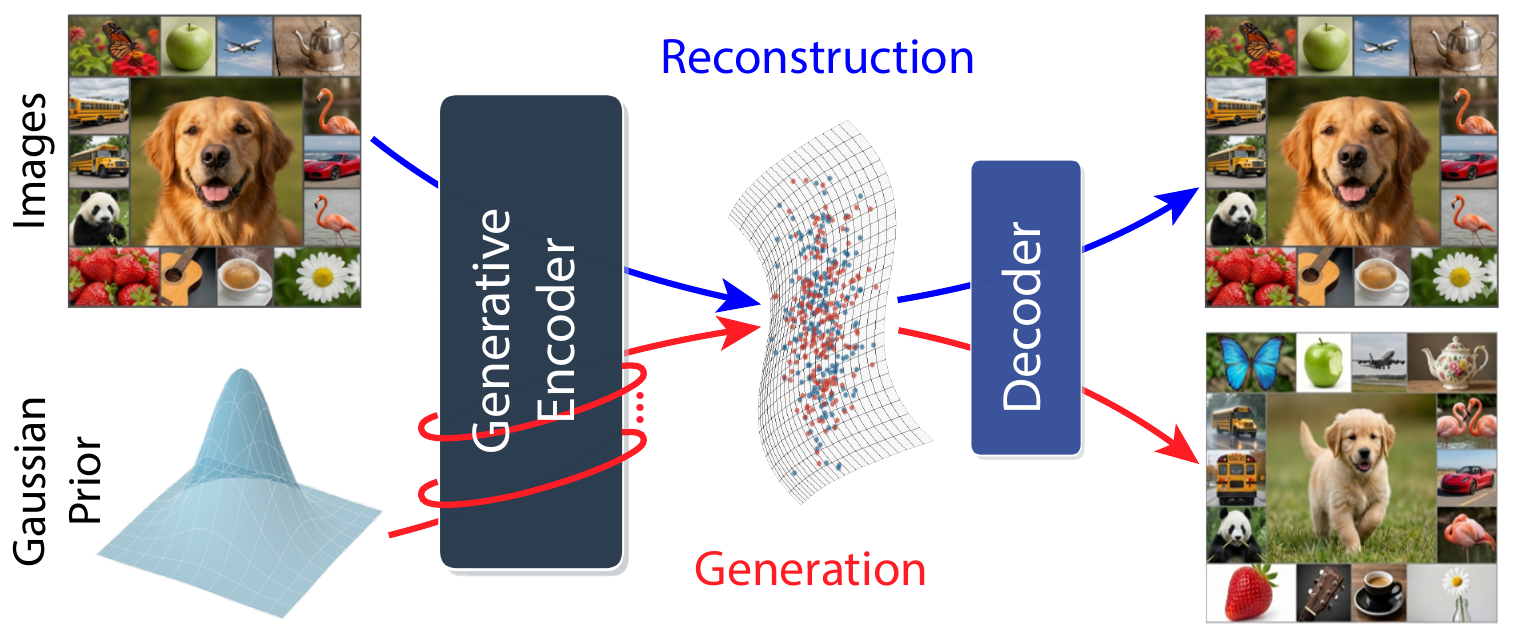}
    \vspace{-4mm}
    \caption{\textbf{Unified Tokenization \& Generation via Generative Encoder (\modelname{}):} We propose a single-stage architecture that unifies tokenization and generation through shared parameters. The \emph{Generative Encoder} operates in two modes: (top) as a tokenizer, it processes image patches and register tokens to produce a latent representation $z_0$; (bottom) as a generator/denoiser, it evolves latents along a flow-matching trajectory to synthesize $z_0$ from Gaussian noise. The latent space is jointly shaped by recon. \& generative objectives from scratch, without external supervision.}
    \label{fig:teaser}
    \vspace{-4mm}
\end{figure}

Modern foundation models~\shamit{\cite{brown2020language, koroteev2021bert, radford2021learning, chen2023pixart, esser2024scaling, polyak2024movie, wan2025wan}}---from language models to video generators, vision--language systems, and scientific generative models---are built around two core operations: tokenization and generation. Tokenization maps high-dimensional observations into a compact latent space that enables both faithful reconstruction and efficient discrimination; generation learns a distribution over this space to synthesize plausible new samples. This division naturally suggests a sequential recipe: first learn a representation space that is easy to reconstruct from and useful for downstream computation; then learn a generative process that samples from that space. As a result, most systems treat tokenization and generation as \emph{separate} design problems \& train them in stages---learning a tokenizer, freezing it \& only then fitting a generator on the induced latent distribution.

This separation is convenient, but it departs from the principle of end-to-end learning and leaves a basic question unresolved: \emph{should tokenization and generation be trained jointly so that each objective can {shape the learned latent space}}? 
In a joint setting, generative pressure could sculpt the latent space toward regions that are easier to model, while reconstruction and inference pressure could preserve instance-specific information and semantic structure. Understanding what emerges when these objectives are trained together---and whether their interaction helps or hurts---is the starting point of this work.

A natural way to pursue this idea is to start from the standard latent generative pipeline. A tokenizer is typically learned as part of an autoencoder with an encoder $E$ and decoder $D$: the encoder maps an image $x$ to a latent sequence $z=E(x)$ and the decoder reconstructs $\hat{x}=D(z)$. A generator then models the latent distribution, most commonly by training a diffusion/flow denoiser~\shamit{\cite{ho2020denoising, song2020score, lipman2022flow}} on noisy versions of $z$: sample a noise level $t$, form $z_t$ by corrupting $z$, and learn a network that predicts the clean latent (or an equivalent parameterization) so that new samples can be generated by starting from noise and iteratively denoising in latent space. In a joint training setting, the same latent $z$ must therefore serve two purposes: it must be decodable by $D$ to preserve instance information, and it must be structured in a way that makes the denoising objective well-posed and easy to learn.

Prior works have explored fully end-to-end training of latent diffusion models by {backpropagating the denoising objective} through the tokenized latents and into the encoder. However, when the tokenizer and diffusion model are optimized primarily through the denoising objective, this can lead to degenerate solutions and poor performance, as observed in REPA-style methods~\shamit{\cite{yu2025repa,leng2025repae}}. To address this, these works propose anchoring the tokenizer with an additional objective that aligns diffusion features to pretrained visual encoders. Although effective, this strategy introduces a third component---a pretrained teacher---to stabilize joint optimization. In contrast, our setting relies \textit{only} on reconstruction and denoising objectives to jointly train the tokenizer and latent generative model, without any external supervision.

\begin{figure}[t]
  \centering
  \includegraphics[width=\columnwidth]{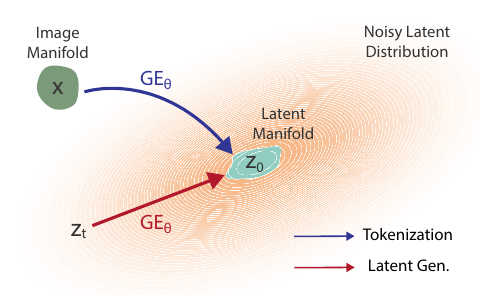}
  \caption{Tokenization and generation can be viewed as the \textit{same latent inference problem under different conditioning regimes.} In tokenization, the full observation $x$ strongly constrains the clean latent, $z_0 \sim p_\theta(z \mid x)$; in generation, a noisy latent $z_t$ provides weaker evidence, and the same target latent $z_0$ is recovered by denoising. {This view motivates using a \textbf{single shared Generative Encoder} ($GE_{\theta}$) for both tokenization and latent denoising.}}
  \label{fig:shared_latent_space}
  \vspace{-3mm}
\end{figure}

{We propose an alternative perspective on end-to-end training.}
Our key insight is that 
tokenization and generation
can be viewed as {the same latent inference problem under different conditioning regimes} 
(see Fig.~\ref{fig:shared_latent_space}). Tokenization can be viewed as a \emph{generative process under strong observability}: given a data point $x$, the model induces a highly concentrated (near-single-point) distribution over latents, yielding a latent $z$ that is consistent with and informative about $x$. Generation corresponds to a {weak-observability regime}, 
where $z$ must be synthesized from noise (and optional conditions) using the learned prior. Under this view, these two operations differ mainly in how much information is available---from the full observation $x$ in tokenization to only a prior in generation. {Motivated by this view, we propose \modelname{}, which jointly trains tokenization and generation end-to-end without external supervision. {\modelname{}} 
ties tokenization \& generation through {a shared-parameter module we call the Generative Encoder (GE)}, 
so that {gradients from both objectives} 
directly shape the same weights, pushing the model toward a representation that is jointly optimal for the two tasks.} Hence the name, \textbf{\modelname{}}: \ul{Uni}fying \ul{T}okenization \& Latent Generation via shared Generative \ul{E}ncoder. See Fig.~\ref{fig:teaser} for an overview.

Concretely, {our system} 
 consists of only two modules: a \emph{Generative Encoder} $\mathrm{GE}_\theta$ and a decoder $\mathrm{D}_\phi$. The GE operates in two modes: (i) \emph{tokenization}, mapping an input $x$ to latent tokens $z=\mathrm{GE}_\theta(x)$, and (ii) \emph{generation}, denoising corrupted latents to produce $\hat z=\mathrm{GE}_\theta(z_t, t)$ at noise level $t$. Thus, the same network serves both as the tokenizer and as the multi-step latent denoiser, with parameters $\theta$ shared across the two objectives.
Training proceeds with two forward passes through $\mathrm{GE}_\theta$. First, we tokenize an input image to obtain clean latents $z$. We then corrupt $z$ using a rectified-flow (flow-matching) process to obtain $z_t$, and pass $z_t$ back through $\mathrm{GE}_\theta$ to predict the corresponding denoising target. The full pipeline is trained end-to-end in a \emph{single stage} by jointly optimizing a pixel-space reconstruction objective and a latent-space flow-matching objective.

{We find that {this end-to-end formulation} 
 yields a strong latent generative model, {with} 
 near state-of-the-art generation and reconstruction fidelity, {while training all modules from scratch rather than relying on large pretrained networks} 
. To {understand} 
 what drives this behavior, {we study other alternatives to end-to-end training in Sec.~\ref{sec:dissecting_generative_encoder}. This includes} an ablation that keeps the full training pipeline fixed but removes parameter tying between the encoder and denoiser. Interestingly, even without explicit weight sharing, the encoder and denoiser exhibit strong per-layer representational alignment, as measured by centered kernel alignment (CKA) \shamit{\cite{kornblith2019similarity}} (See Fig.~\ref{fig:cka_analysis_final}), suggesting that tokenization and denoising are intrinsically compatible tasks in our setting. Further analysis (see Sec.~\ref{sec:dissecting_generative_encoder}) indicates that the model differentiates the two modes primarily through normalization: the tokenization and denoising pathways occupy different norm/scale regimes, while attention and MLP sublayers remain highly reusable across both. In fact, recent concurrent work on {Unified Latent} \cite{heek2026unifiedlatentsultrain} investigates a closely related two-module formulation. It can be interpreted as a special case of our end-to-end setting, aligning closely with our separate-weights ablation. While this separate-weights variant performs {almost as competitively,} 
 we find that {parameter tying} 
 yields the best overall rFID/gFID {trade-off} 
 in our experiments (Fig.~\ref{fig:weight_sharing_analysis}).
 Overall, these results provide a concrete single-stage recipe in which reconstruction and denoising objectives can jointly shape the latent space, rather than being optimized in disjoint stages. Practically, this means \textbf{one training job and one model} to store and update, while retaining a near-SOTA tokenizer and generator.}

\section{Related Work}
\label{sec:related_work}


\paragraph{Tokenization \& Generation via Auto-Encoding:} Variational autoencoders (VAEs) \cite{kingma2014vae} introduced a principled framework for learning probabilistic latent representations while enabling generation through a simple Gaussian prior. This foundational work established that reconstruction and generation can be learned within a single model, though the Gaussian prior and likelihood assumptions often limit sample quality. Extensions such as VQ-VAE \cite{vandenoord2017vqvae} and VQ-GAN \cite{esser2021vqgan} improved representation learning by introducing discrete latent spaces and adversarial training, respectively; in practice, many widely used autoencoder-based tokenizers for diffusion are trained with GAN-style \shamit{\cite{goodfellow2020generative}} losses. However, in modern latent diffusion pipelines \cite{peebles2023dit, ma2024sit}, these VAE/VQGAN-style models primarily serve as \emph{tokenizers} for a downstream diffusion model trained in the resulting frozen latent space; their standalone generative capability is typically weaker and is therefore rarely used in practice. In standard downstream diffusion training, the denoising/generative gradients never flow back into the tokenization process, preventing the representation from being shaped by the needs of generation. To address this, we couple tokenization with a latent denoising objective and train a single model end-to-end, allowing the encoder to be directly shaped by generative learning. For 
simplicity, we eliminate adversarial losses in all experiments unless otherwise stated.

\vspace{-2mm}
\paragraph{Self-supervised Visual Encoders for Generation:} Recent advances in self-supervised learning have produced powerful visual encoders that go beyond naive reconstruction objectives. Masked autoencoders (MAE) \cite{he2022mae} show that reconstructing masked patches can learn strong visual representations at scale. DINO-style models \cite{caron2021dino, oquab2023dinov2} learn semantic features via self-distillation without labels, yielding representations that capture both local and global image structure. Building on these encoders, recent methods such as REPA \cite{yu2025repa}, REPA-E \cite{leng2025repae}, and RAE \cite{rae2025diffusion} leverage pretrained SSL models as extra supervision for diffusion model training. REPA improves training efficiency and sample quality by aligning intermediate diffusion features with SSL representations. REPA-E extends this idea by jointly tuning the VAE and diffusion model to better match the SSL space, while RAE replaces the VAE encoder with an SSL encoder and trains a separate decoder in a subsequent stage for reconstruction. While these approaches achieve strong generation quality, they further increase pipeline staging and do not study how reconstruction and generation can jointly shape shared model parameters. In contrast, we focus on a single-stage training approach that learns tokenization and generation jointly, without access to pretrained SSL encoders.


\vspace{-2mm}
\paragraph{Pixel-Space Diffusion Models:} Pixel-space diffusion models denoise directly in the RGB domain, avoiding a learned latent space but facing sharper scaling issues at high resolution. As resolution increases, stronger local redundancy lets fixed noise be averaged out, raising effective SNR and making denoising too easy; thus prior work scales noise (or reweights the loss) to keep difficulty/SNR consistent across resolutions \cite{hoogeboom2023simple,chen2023importance,kingma2023understanding}. This motivates architectural adaptations tailored to high-resolution pixel modeling: SiD2 \cite{hoogeboom2024simpler} trims U-Net skip connections and reduces high-resolution feature capacity; PixelFlow \cite{chen2025pixelflow} alternates denoising with progressive upsampling; and methods such as PixNerd \cite{wang2025pixnerd}, PixelDiT \cite{yu2025pixeldit}, and DiP \cite{chen2025dip} introduce specialized heads to better handle fine-grained inputs. JiT \cite{li2025jit} takes a complementary minimalist stance, training a plain ViT generator directly on raw patches without tokenizers, pretraining, or auxiliary losses. Unlike these pixel-space approaches that focus on learning a generator, we study both latent-space inference (via tokenization) and generation.


\vspace{-2mm}
\paragraph{Concurrent works:}
{Several concurrent papers} 
 have explored closely related directions toward unifying tokenization and latent generative modeling. The closest to our setting is Google’s Unified Latents~\cite{heek2026unifiedlatentsultrain}, which studies end-to-end training of a tokenizer together with a latent generator and is closely aligned with our separate-weights ablation (i.e., an encoder and denoiser trained jointly without parameter sharing). In contrast to our single-stage results, their strongest numbers rely on an additional second-stage diffusion fine-tuning step (see Appendix B of \citealt{heek2026unifiedlatentsultrain}). Latent Forcing~\cite{baade2026latentforcingreorderingdiffusion} extends pixel-space diffusion (JiT) by denoising pretrained DINO latents alongside image patches through a shared bottleneck, but does not learn the latent space from scratch. Another concurrent effort~\cite{CheferEsser2026selfflow} adds an auxiliary self-supervised objective alongside the diffusion/flow objective, but similarly operates in a latent space defined by a pretrained encoder rather than jointly learning the tokenizer and generator end-to-end. In contrast to these works, our primary emphasis is on understanding the capabilities of a single-stage, end-to-end trained latent diffusion model. To this end, we study a perspective in which encoding \& denoising are performed by the same network parameters.

\section{Unifying Tokenization \& Latent Denoising}



Can we jointly train a tokenizer and a generator end-to-end in a single stage, such that gradients from one objective meaningfully shape the other? 
{The answer is yes: we show that single-stage end-to-end training can learn a latent space that supports both high-fidelity reconstruction and iterative generation.} While recent work has begun to explore end-to-end training, most approaches still rely on multi-stage pipelines (e.g., pretraining or freezing parts of the system) or introduce external supervision from pretrained representation models. These design choices can be effective, but they make it harder to isolate and study the intrinsic interaction between tokenization and generation.
In this work, we take a step toward single-stage joint tokenization and generation without external supervision, using a single unified network trained simultaneously with reconstruction and latent denoising objectives. 



In many ways, an early and elegant solution to this already exists: \emph{variational autoencoders} (VAEs) jointly learn an encoder--decoder for reconstruction while also imposing a simple latent prior, typically $\mathcal{N}(0,I)$, that enables sampling and generation. This classical design suggests that tokenization \& generation need not be separated into distinct stages.


\subsection{From VAE to \modelname{}}
In a VAE, the ``tokenizer'' is the encoder, $\mathrm{E_\theta}$: it maps a data point $x$ to a conditional latent distribution $q(z\mid x)$ rather than a single code. The decoder, $\mathrm{D_\psi}$, reconstructs by sampling $z \sim q(z\mid x)$ and mapping back to data space via $p(x\mid z)$. For any generative model, a central requirement is to map an easy-to-sample distribution into an expressive latent space that supports high-quality decoding. VAEs meet this requirement by regularizing the encoder so that its latent distribution remains close to a simple prior $p(z)=\mathcal{N}(0,I)$ (through the KL loss term), making generation as simple as sampling $z \sim p(z)$ and decoding.
\vspace{-1mm}
\[\text{\textbf{VAE:}} \hspace{2mm} z = E_\theta(x); \hat{x} = D_\psi(z); \]
\vspace{-7mm}

Notably, VAE-family encoder--decoder tokenizers have become a standard building block in modern vision and video foundation-model pipelines: high-dimensional visual inputs are first compressed into latents via a VAE/VQ-style encoder, but generation is performed in latent space by a separate model. In this regime, these autoencoders function primarily as tokenizers rather than as the final generative model, since a simple Gaussian prior typically does not reach the sample fidelity of modern diffusion generators.

Modern high-fidelity generative models therefore replace VAE-style Gaussian prior sampling with a learned iterative generative process, while retaining the VAE's role as the tokenizer. In latent diffusion and flow models, a VAE-style encoder first maps data into a compact latent space, and a separate denoising model, $\mathrm{G_\phi}$, is trained to transform Gaussian noise into samples from the latent data distribution via iterative denoising. In practice, this is often implemented as a staged pipeline: the tokenizer is trained and frozen; the denoiser is trained on top of the fixed latent space.
\vspace{-1mm}
\[\text{\textbf{LDM:}} \hspace{2mm} z = E_\theta(x); \hat{x} = D_\psi(z); \hat{z} = G_\phi(z_t, t); \]
\vspace{-6mm}


\begin{figure}[t]
    \centering
    \includegraphics[width=\columnwidth]{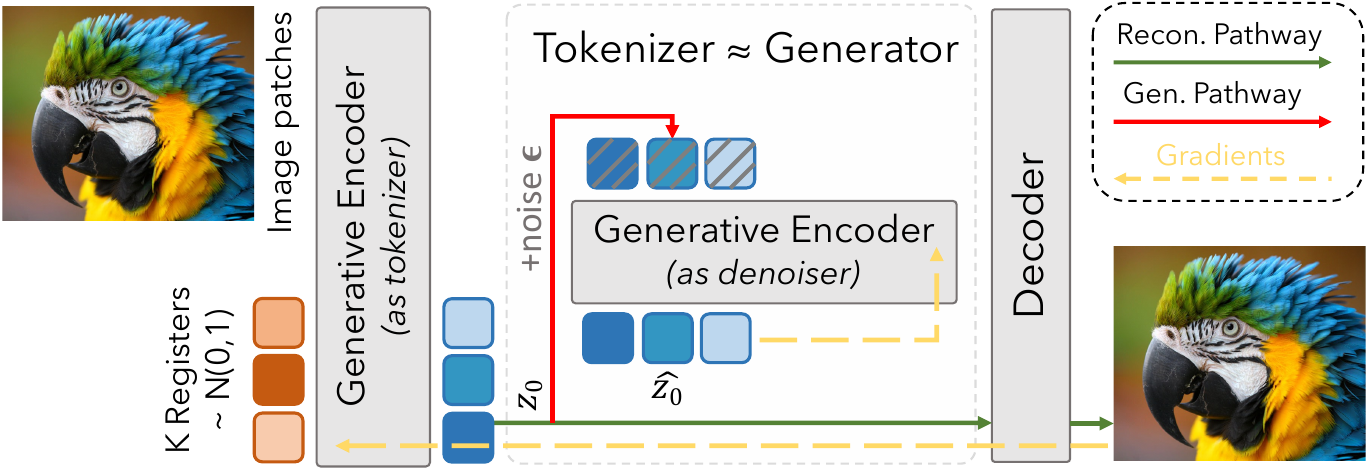}
    \vspace{0.5mm}
    \caption{\textbf{\modelname{} Training Pipeline} uses two forward passes through the Generative Encoder: first, mapping (distilling) image patches into latent registers, and second, denoising a noised version of those latents, with weights shared across both passes. Training combines reconstruction losses with a denoising loss $|\tilde{\hat{z}}_0 - sg(\tilde{z}_0)|$.}
    \label{fig:architecture}
\end{figure}

{\modelname{} replaces the separate tokenizer and latent denoiser with a shared set of parameters, the Generative Encoder, as demonstrated in Fig.~\ref{fig:shared_latent_space}.
This shared module} 
retains the simplicity of the autoencoder interface---an encoder and a decoder---while enabling \emph{single-stage} learning of both tokenization and generation. {Paired with a decoder $D_\psi$ that maps latents back to image space,} 
the Generative Encoder $\mathrm{GE}_\theta$ operates in two modes. In \emph{tokenization mode}, $\mathrm{GE}_\theta$ maps an image $x$ to latent tokens $z=\mathrm{GE}_\theta(x)$ optimized for reconstruction, without enforcing an explicit KL-to-Gaussian bottleneck. In \emph{generation mode}, the same $\mathrm{GE}_\theta$ is used as a latent denoiser: given a noisy latent $z_t$ and noise level $t$, it predicts the corresponding denoising target, enabling iterative sampling from Gaussian noise at inference time. {Sharing parameters across these two modes lets gradients from both objectives jointly shape the same weights in a single training job}. 
This yields a minimal end-to-end pipeline with performance approaching modern latent generative models, with the resulting  formulation as:
\vspace{-1mm}
\[\text{\textbf{\modelname{}:}} \hspace{2mm} z = GE_\theta(x); \hat{x} = D_\psi(z); \hat{z} = GE_\theta(z_t, t); \]
\vspace{-7mm}

\subsection{End-to-End Training for \modelname{}}

\textbf{Training Pipeline:}
We adopt a Vision Transformer (ViT) \shamit{\cite{dosovitskiy2020image}} backbone for both the generative encoder and the decoder, motivated by the strong empirical performance of Transformer architectures in diffusion/flow denoising. The {generative encoder} $\mathrm{GE}_\theta$ must support two operating modes with compatible input/output types: a {\emph{tokenization pathway}}, which ingests image patch tokens and produces a compact latent representation, and a {\emph{generation (denoising) pathway}}, which ingests noisy latents along a flow or diffusion trajectory that connects the latent distribution to a standard normal prior.

To unify the input format across pathways, we represent the latent $z$ as a fixed set of $K$ \emph{register tokens}. In the tokenization pathway, we concatenate the image patch tokens with $K$ registers, initializing the registers as i.i.d.\ Gaussian noise, $\mathcal{N}(0,I)$, to match the input distribution at the maximum noise level. The concatenated sequence is processed with self-attention in a \emph{first} forward pass through $\mathrm{GE}_\theta$. We then discard the patch tokens and retain only the updated registers. These updated registers serve as the image latents $z_0$, having absorbed the relevant information from patches through attention. The decoder $D_\psi$ consumes $z_0$ and reconstructs the image using a ViT-style stack followed by a lightweight unpatchification head to produce pixels.

In the generation (denoising) pathway, we first corrupt the clean latents $z_0$ to obtain a noisy latent $z_t$ at noise level $t$ (using our rectified-flow / flow-matching corruption process) and then use $z_t$ to initialize the same $K$ registers. No image patches are concatenated in this pathway. A \textit{second} forward pass through $\mathrm{GE}_\theta$ (now in generation mode, conditioned on $t$ and optional class information) predicts the denoising target; in our implementation we use $x$-start prediction, i.e., $\hat z_0=\mathrm{GE}_\theta(z_t,t)$, so that the denoiser output lies in the same space as the tokenization output. To avoid degenerate solutions where the denoiser objective collapses the latent space, we stop gradients through the clean latents used to form $z_t$ (i.e., we detach $z_0$ before noising).

The final layer of $\mathrm{GE}_\theta$ is a normalization module. Empirically, we find that LayerNorm \shamit{\cite{ba2016layer}} with learnable scale and shift parameters performs best. As a result, the clean latents $z_0$ (from the tokenization pathway), the denoised predictions $\hat{z}_0$ (from the generation pathway), and the model outputs at each denoising step during inference are all normalized. 


Overall, each training iteration performs two forward passes through the shared $\mathrm{GE}_\theta$: an image-conditioned pass to produce clean latents for reconstruction, followed by a latent-only pass to denoise a corrupted version of those latents. The full system is trained end-to-end in a \emph{single stage} by jointly optimizing a pixel-space reconstruction objective (via $D_\psi$) and a latent-space denoising objective (via $\mathrm{GE}_\theta$).

\begin{figure}[t]
  \centering
  \includegraphics[width=0.9\columnwidth]{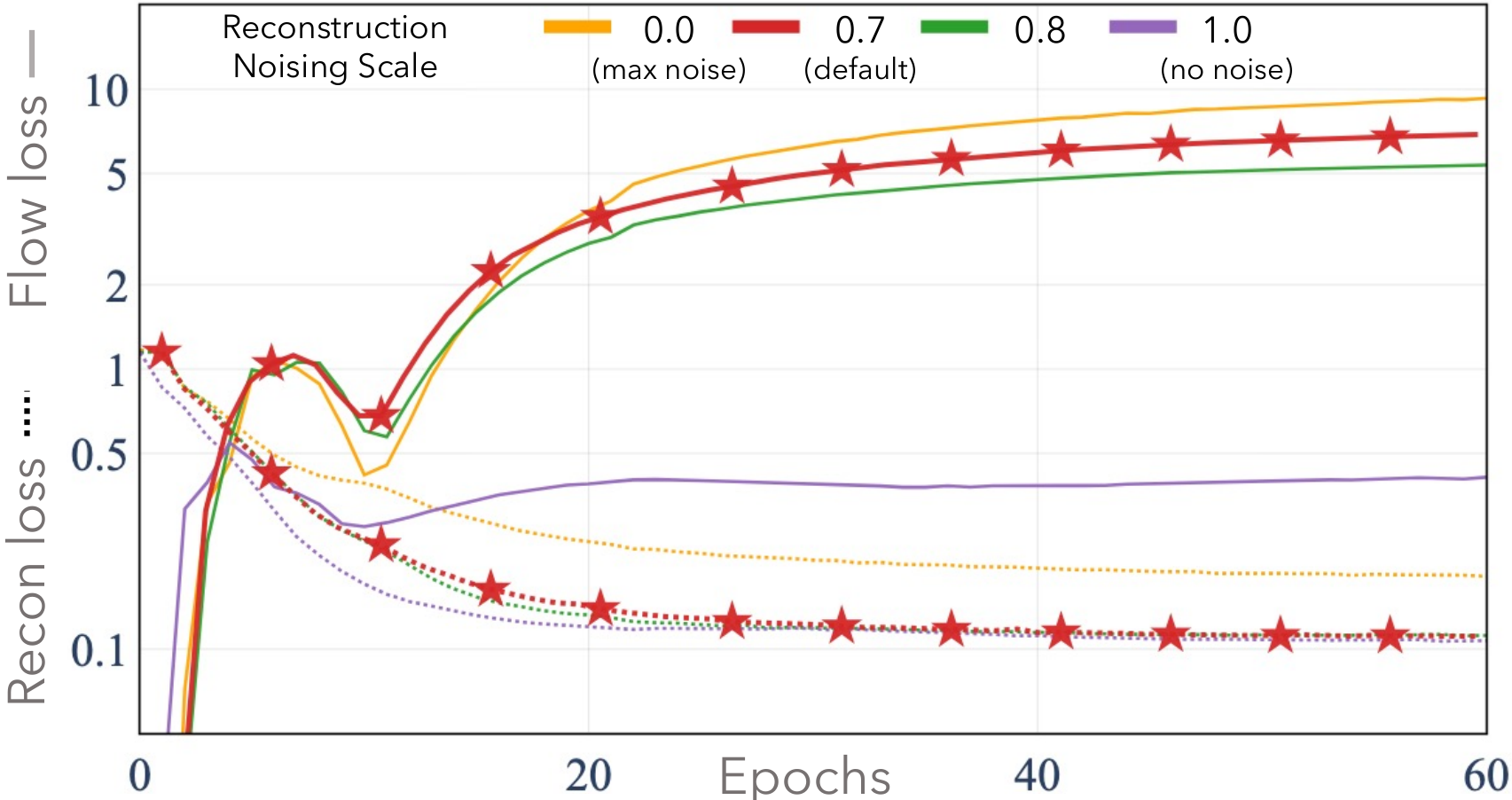}
  \caption{\textbf{\modelname{}'s Training dynamics:} The conflicting nature of the reconstruction and denoising objectives leads to an adversarial training behavior when trained jointly. The dotted lines (zoomed in) represent different ablations {(see Appendix)} based on the scale of noise added in reconstruction pathway for decoder robustness.}
  \vspace{-4mm}
  \label{fig:training_dynamics}
\end{figure}

\textbf{Training Objectives:}
We optimize two losses computed from the two forward passes described above.
For reconstruction, we encode the image into clean latents $z_0=\mathrm{GE}_\theta(x)$, inject small Gaussian noise $\tilde{z}_0 = z_0 + \sigma\epsilon$ with reconstruction noise scale $\sigma=0.7$ following~\citet{leng2025repae,yu2025repa}, and decode $\hat{x}=D_\psi(\tilde{z}_0)$.
The reconstruction loss combines pixel-level and perceptual terms: $\mathcal{L}_{\text{recon}} = \|\hat{x}-x\|_1 + \text{LPIPS}(\hat{x},x)$.
For generation, we apply rectified flow matching~\citep{liu2023rectified} on the latents.
Given clean latents $z_0$, we construct noisy latents $z_t = t z_0 + (1-t) \epsilon$ with $\epsilon \sim \mathcal{N}(0,I)$ and $t \sim \mathcal{U}[0,1]$ (where $t{=}1$ corresponds to clean data and $t{=}0$ to pure noise), then train the generative encoder to predict clean latents via $\hat{z}_0 = \mathrm{GE}_\theta(z_t,t)$.
We minimize $\mathcal{L}_{\text{flow}} = \mathbb{E}_{t,\epsilon}[\|\hat{z}_0-\text{sg}(z_0)\|_2^2]$, where $\text{sg}(\cdot)$ denotes stop-gradient to prevent degenerate solutions.
The total objective is the sum of reconstruction and generation losses.

\textbf{Inference:} At inference, the Generative Encoder can serve as the tokenizer by mapping an input image to its latent representation in a single forward pass. For generation, we start from a class label and noisy latent registers, and iteratively refine them through multiple passes of the GE into clean, decodable latents (shown as red loops in Fig.~\ref{fig:teaser}).

\subsection{Understanding \modelname{}'s Training Dynamics}

\paragraph{The adversarial nature of joint training.}
Jointly training tokenization and generation under weight sharing induces non-trivial dynamics. In a standard LDM, the latent space is produced by a pretrained (and typically frozen) tokenizer, so the generative objective does not shape the latent interface. In \modelname{}, reconstruction and generative objectives are optimized jointly over the same parameters, so each objective can influence the representations used by the other.

This dynamic is best understood as the search for a latent space that satisfies two distinct pressures shaping its structure.
The reconstruction objective drives the encoder to maximize information content, preventing the latent representation from becoming too coarse to capture instance-specific detail.
Simultaneously, the generative objective constrains how this information is encoded: it penalizes learning fragile representations whose semantic content can be easily destroyed by noise, since such instability makes denoising harder.
Consequently, joint optimization balances these pressures, finding a latent space that is rich enough for reconstruction yet robust enough against perturbations.
By forcing the encoder to adopt this robust geometry, the generative loss effectively molds the latent space into one that is intrinsically easier to denoise---facilitating high-fidelity generation.


Empirically, this interaction can resemble an ``adversarial'' game: the two losses do not necessarily decrease monotonically together.
Improvements in generative fidelity can even coincide with an increase in denoising loss, as shown in Fig.~\ref{fig:training_dynamics} (see red curves with star markers).
Crucially, a rising denoising loss does not imply worse generation.
Instead, it often signals that the latent space is becoming richer and more informative to satisfy the reconstruction objective, making the denoising task harder but the resulting samples more realistic.
During training, we often observe generation metrics (e.g., FID/IS) improving even as the denoising loss increases, until the system reaches a stable equilibrium.
Similar to GAN-style training, the goal is therefore not to drive all losses to zero, but to reach \emph{stable} training dynamics where the latent space balances information density with generative robustness.
This perspective is also consistent with modern diffusion/flow models, where the denoising loss typically stabilizes at a non-zero value.





\section{Analyzing \modelname{}'s Generative Encoder}
\label{sec:dissecting_generative_encoder}

In \modelname{}, we pursue end-to-end training by \emph{sharing} parameters between the encoder and denoiser roles of a single network. This choice suggests a natural hypothesis: parameter tying encourages the model to develop a common latent ``language''---shared internal features and transformations that simultaneously support reconstruction and iterative denoising-based sampling.

To better understand this design choice, we study two alternative routes to end-to-end latent diffusion training that each relax a component of our Generative Encoder mechanism. First, we remove parameter tying, maintaining \textit{separate} encoder and denoiser networks while still training both objectives jointly. Second, we remove the stop-gradient through clean latents, allowing {denoising gradients to backpropagate into the tokenization pathway}. Together, these alternatives help isolate the role of weight sharing and gradient flow in our end-to-end
formulation. {Finally, we also study these end-to-end training approaches through the lenses of representation alignment and compression.}

\begin{figure}[t]
  \centering
  \includegraphics[width=0.94\columnwidth]{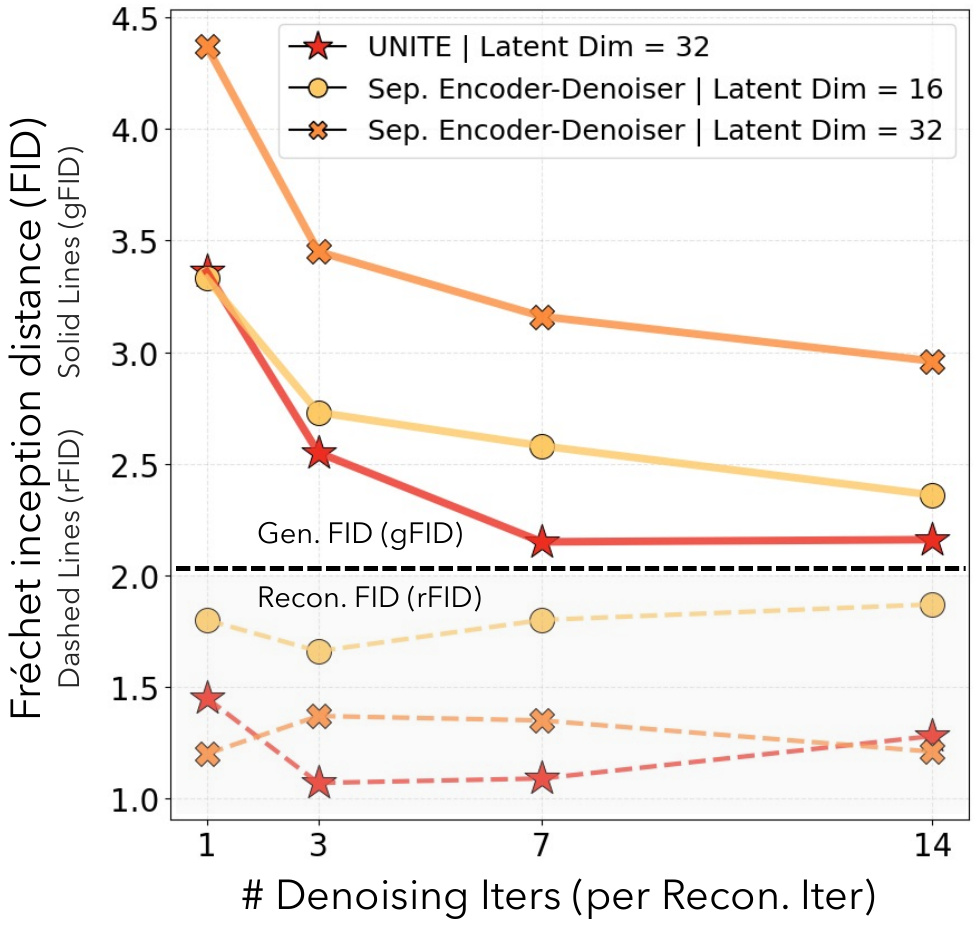}
  \caption{\textbf{Weight-shared vs.\ Separate Enc-Denoiser training.} UNITE uses a single Generative Encoder, sharing weights between tokenization and generation. To isolate the effect of weight sharing, we keep the rest of the end-to-end training pipeline fixed, including the stop-gradient that prevents denoising gradients from flowing into the tokenized output. Both UNITE and the separate encoder-denoiser ablation attain \textit{competitive} performance, with {UNITE benefiting from more denoising-to-reconstruction steps ratio} during training, achieving the best overall rFID--gFID trade-off.}
  \label{fig:weight_sharing_analysis}
  \vspace{-3mm}
\end{figure}

\begin{figure*}[t]
  \centering
  \includegraphics[width=\textwidth]{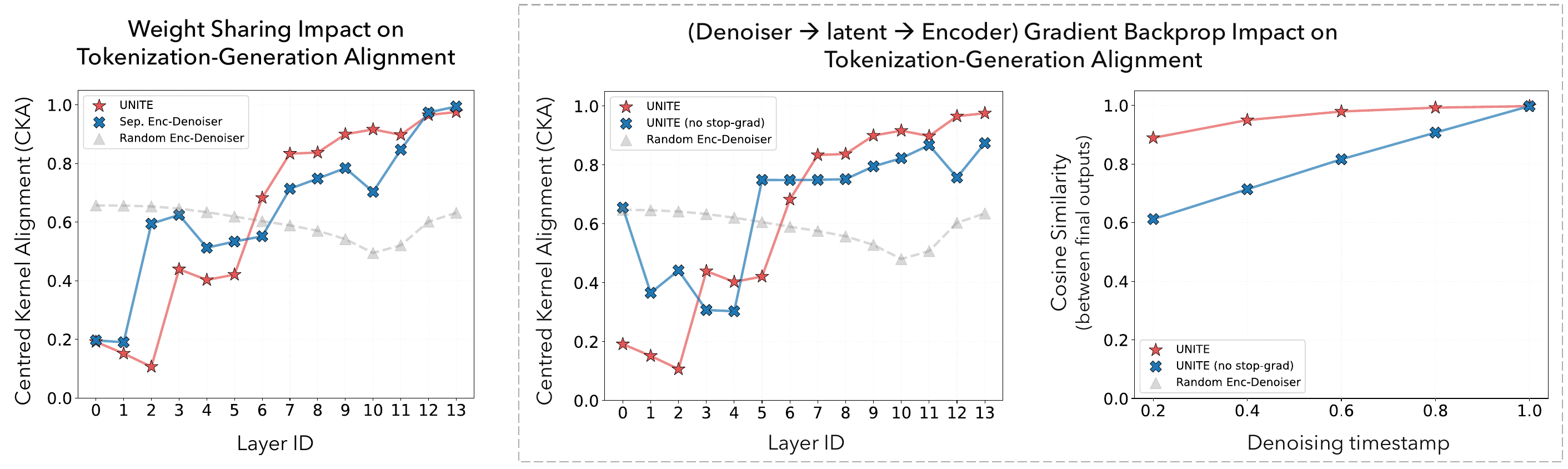}
  \caption{\textbf{Representation alignment between tokenization and generation pathways.} We measure alignment between tokenization and denoising activations using CKA and cosine similarity. Given an input image, we first record intermediate activations along the tokenization pathway, then corrupt the encoded latent and record the corresponding denoising-pathway activations. \textbf{Left:} both the weight-shared UNITE model and the separate encoder--denoiser ablation exhibit strong alignment, especially in later layers, indicating that \textit{tokenization and denoising are intrinsically aligned tasks}. \textbf{Middle:} removing the stop-gradient and backpropagating denoising gradients through the latent \textit{weakens late-layer alignment}, even though the denoising objective still matches the final latent target. \textbf{Right:} cosine similarity on the final latents decreases at lower denoising timesteps in the no-stop-gradient setting, suggesting that direct gradient backpropagation from denoising into tokenization leads to a less cleanly shared representation (see Fig.~\ref{fig:denoising_analysis} for visual interpretation).}
  \label{fig:cka_analysis_final}
\end{figure*}

\begin{figure*}[t]
  \centering
  \includegraphics[width=\textwidth]{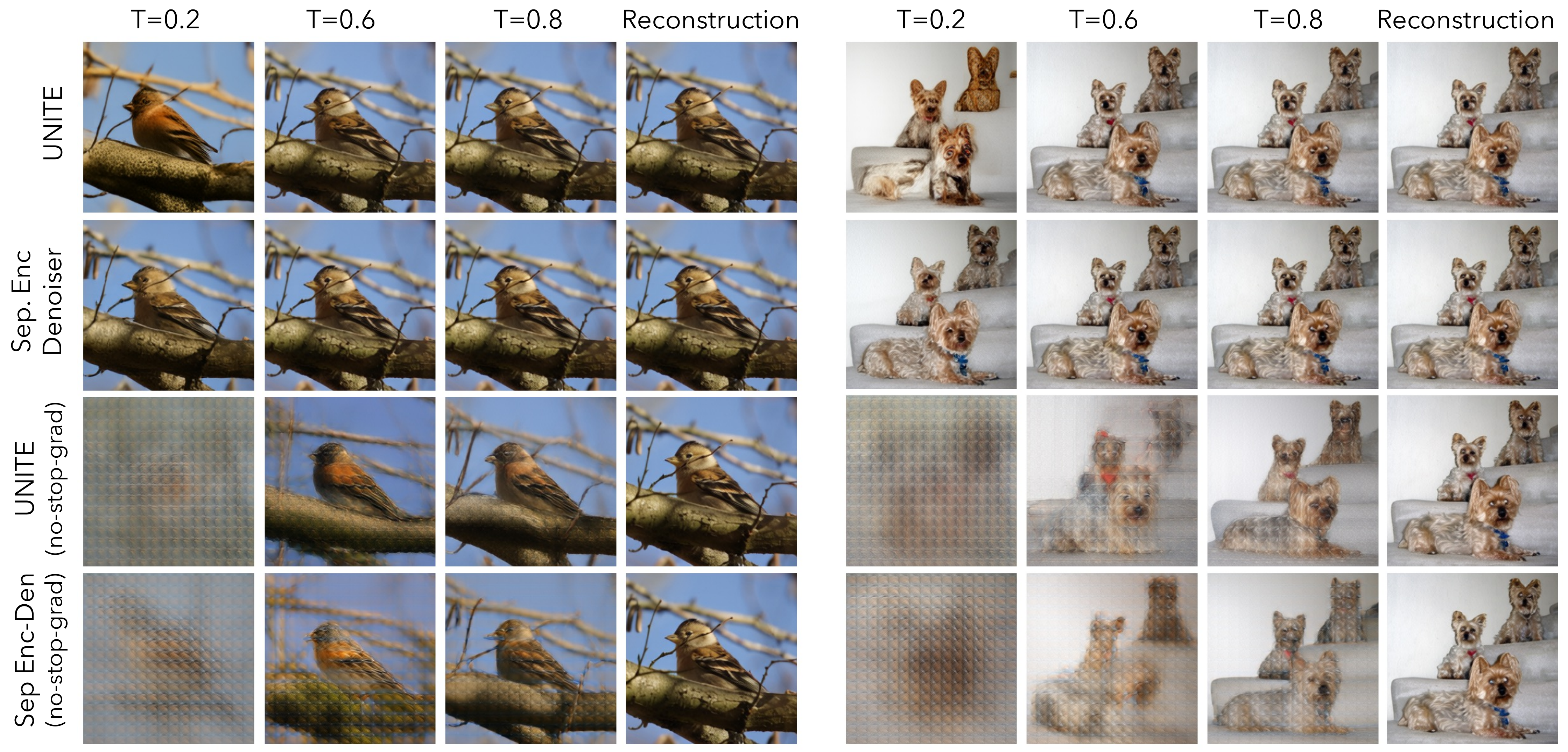}
  \vspace{-1mm}
  \caption{\textbf{Analyzing the denoising trajectory.} Given an input image, we first encode it into latents, corrupt the latent with noise, and then decode the denoised prediction at different noise levels (first three columns). The final column shows direct decoding of the clean latent. Although all four models achieve competitive aggregate rFID/gFID, the stop-gradient variants (first two rows)--UNITE and the separate encoder-denoiser ablation--exhibit markedly cleaner intermediate denoising trajectories, with higher PSNR to the input image across all noise levels. This result is consistent with the representation-alignment in Fig.~\ref{fig:cka_analysis_final}, which shows drop in alignment at final layers.}
  \label{fig:denoising_analysis}
  \vspace{-1mm}
\end{figure*}

\paragraph{Weight-Shared vs. Separate Encoder–Denoiser Training:}
Our Generative Encoder ties the encoding and denoising roles by sharing parameters. As an ablation, we keep the entire end-to-end pipeline fixed---including the stop-gradient that prevents denoising gradients from flowing through the tokenization output into the encoder--but instantiate \emph{separate} networks for the encoder and the denoiser. {In this separate-networks ablation, the encoder \& denoiser are optimized for their own objectives, with no gradient interaction involved.} 

If the weight-shared Generative Encoder matches or improves upon this separate-weights variant, it already offers a practical advantage: fewer parameters to store and update, and a shorter description length (MDL) for the learned model. Fig.~\ref{fig:weight_sharing_analysis} shows that, while the separate-weights ablation is competitive, parameter tying yields the best overall reconstruction--generation trade-off. Specifically, we report rFID and gFID as a function of the number of denoising (flow) steps performed per reconstruction step during training. Under weight sharing, increasing the number of flow steps consistently improves generation fidelity, reducing gFID from 3.33 to 2.12 as the number of flow iterations is increased by $14\times$. This indicates that the latent space becomes more sampleable--while maintaining, or slightly improving, reconstruction fidelity, suggesting that the representation also remains information-preserving at the chosen compression dimension. Next, we study the role of {the} stop gradients operator between the denoiser and the tokenizer.

\vspace{-2mm}
\paragraph{Backpropagating Denoising Gradients through the Encoder:}
Throughout this work, we stop denoising gradients from flowing through the clean latent into the tokenization pathway. Concretely, after the tokenization pass produces $z_0=\mathrm{GE}_\theta(x)$, we apply $\mathrm{sg}(\cdot)$ before constructing the noised latent $z_t$ used in the denoising pass. As a result, the flow-matching objective updates $\mathrm{GE}_\theta$ only through the second (denoising) forward pass, rather than also directly shaping tokenization through gradients flowing into $z_0$.

Importantly, this does \emph{not} decouple tokenization and generation: in the weight-shared Generative Encoder, reconstruction and denoising still act on the same set of network parameters, so both objectives jointly shape the learned representation. The stop-gradient only removes the more direct route in which denoising gradients also flow through the clean latent itself. In the separate encoder--denoiser setting, removing this stop-gradient yields a two-network end-to-end regime closely analogous to concurrent work on {Unified Latents} (UL)~\cite{heek2026unifiedlatentsultrain}, which jointly trains separate encoder and denoiser modules without parameter sharing. We therefore study what happens when denoising gradients are allowed to backpropagate through the clean latent {(termed the no-stop-grad setting in the following paragraphs)}, both in our weight-shared GE setting and in the separate encoder--denoiser ablation. 


{Looking at rFID/gFID, removing the stop-gradient improves the separate encoder--denoiser ablation from $2.60/1.30$ to $2.24/0.85$ (gFID/rFID), indicating that {end-to-end joint training of tokenization and generation is promising}. As {noted} in the concurrent Unified Latents~\cite{heek2026unifiedlatentsultrain} {(their Appendix B)}, obtaining the best performance in the no-stop-gradient setting requires tuning the denoising-to-reconstruction loss ratio. By contrast, for UNITE, we obtain the best performance (gFID $= 2.12$, rFID $= 1.1$) with stop-gradient in place. One possible hypothesis is that, under weight sharing, the two objectives already interact through a common parameter set, so allowing denoising gradients to additionally flow through the clean latent introduces extra (asymmetric) gradient interference. In this sense, weight sharing itself acts as a natural coupling mechanism between the two tasks: simply increasing the number of flow iterations improves performance, without requiring as much loss-weight tuning.}
We now showcase representation alignment and compression-based analysis.

\begin{table}[t]
\centering
\caption{\textbf{ImageNet 256$\times$256 generation.} Our approach outperforms both recent single-stage pixel baselines and standard two-stage latent diffusion frameworks by a large margin.}
\label{tab:imagenet_results}
\scalebox{0.85}{
\begin{tabular}{lcccc}
\toprule
Method & \makecell{Aux.\\Token} & Params & FID$\downarrow$ & IS$\uparrow$ \\
\midrule
\rowcolor{gray!15} \multicolumn{5}{l}{{\textit{Single-stage Frameworks}}} \\
JiT-B/16 & - & 131M & 3.66 & 275.1 \\
\textbf{\modelname{}-B (Ours)} & \textbf{Joint} & 217M & \textbf{2.12} & \textbf{294.1} \\
\hdashline
\addlinespace[3pt]
RIN & - & 410M & 3.42 & 182.0 \\
JiT-L/16 & - & 459M & 2.36 & 298.5 \\
ADM-G & - & 554M & 4.59 & 186.7 \\
\textbf{\modelname{}-L (Ours)} & \textbf{Joint} & 589M & \textbf{1.73} & \textbf{296.0} \\
\hdashline
\addlinespace[3pt]
PixelFlow-XL/4 & - & 677M & 1.98 & 282.1 \\
PixNerd-XL/16 & - & 700M & 2.15 & 297 \\
\textbf{\modelname{}-XL (Ours)} & \textbf{Joint} & 806M & \textbf{1.75} & \textbf{309.9} \\
JiT-H/16 & - & 953M & 1.86 & 303.4 \\
SiD & - & 2B & 2.44 & 256.3 \\
VDM++ & - & 2B & 2.12 & 267.7 \\
JiT-G/16 & - & 2B & 1.82 & 292.6 \\
\midrule
\rowcolor{gray!15} \multicolumn{5}{l}{{\textit{Two-stage Frameworks}}} \\
DiT-XL/2 & SD-VAE & 675M+49M & 2.27 & 278.2 \\
SiT-XL/2 & SD-VAE & 675M+49M & 2.06 & 277.5 \\
\addlinespace[3pt]
\midrule
\rowcolor{gray!15} \multicolumn{5}{l}{{\textit{Two-stage Frameworks with Aux Supervision (DINOv2)}}} \\ REPA-B & SD-VAE & 130M+49M & 2.15 & 268.3 \\ RAE-B & RAE-tok & 130M+415M & 2.08 & 275.1 \\ \addlinespace[3pt] REPA-SiT-XL/2 & SD-VAE & 675M+49M & 1.42 & 305.7 \\ LightningDiT-XL/2 & VA-VAE & 675M+49M & 1.35 & 295.3 \\ DDT-XL/2 & SD-VAE & 675M+49M & 1.26 & 310.6 \\ RAE-DiT$^\text{DH}$-XL/2 & RAE & 839M+415M & 1.13 & 262.6 \\
\midrule
\rowcolor{gray!15} \multicolumn{5}{l}{\textit{Concurrent works}} \\
LF-DiT-L & DINOv2 & 465M & 2.48 & — \\
\bottomrule
\end{tabular}
}
\vspace{-6mm}
\end{table}

\vspace{-3mm}
\paragraph{Tokenization-Generation Representation Alignment Analysis:} As shown in Fig.~\ref{fig:shared_latent_space}, tokenization can be viewed as a generative process under strong observability, $p_\theta(z \mid x)$, whereas generation corresponds to unconditional sampling from the induced prior, $z \sim p_\theta(z)$. This viewpoint suggests that the two tasks may be aligned, and motivates measuring representational alignment between the two modes. We test this by measuring alignment between tokenization-pathway \& denoising-pathway activations using Centered Kernel Alignment (CKA / CKNNA) and Cosine Similarity (Fig.~\ref{fig:cka_analysis_final}).

Several aspects of our design encourage alignment. First, both modes are trained to operate in the same latent space: the denoiser is supervised to predict the corresponding clean latent for a corrupted version of the encoded latent. Second, the GE receives the same latent register parameterization in both modes; during tokenization these registers are initialized from $\mathcal{N}(0,1)$, reducing input-domain mismatch between tokenization and generation. Finally, we adopt architectural and optimization choices that limit drift between modes: (i) consistent normalization throughout the network (within blocks and at the encoder output), (ii) matched conditioning interfaces across modes (e.g., time and class signals injected in analogous ways) to avoid mode-specific shortcuts, (iii) conservative optimization (learning-rate warmup and schedules) to prevent one objective from dominating shared parameters early in training.


\begin{table}[t]
\centering
\caption{\textbf{ImageNet 256$\times$256 reconstruction.} Our tokenizer achieves competitive rFID without adversarial loss (Adv.) or pretrained encoders. All \modelname{} rows use base backbone at 120 eps.}
\label{tab:reconstruction}
\scalebox{0.85}{
\begin{tabular}{lccc}
\toprule
Tokenizer & Adv. & Pretrained Encoder & rFID$\downarrow$ \\
\midrule
\rowcolor{gray!15} \multicolumn{4}{l}{\textit{{With adversarial / external supervision}}} \\
SD-VAE & \checkmark & - & 0.62 \\
DC-AE-f32 & \checkmark & - & 0.69 \\
RAE & \checkmark & DINOv2 & 0.58 \\
VA-VAE & \checkmark & DINOv2 & \textbf{0.28} \\
\bottomrule
\rowcolor{gray!15} \multicolumn{4}{l}{\textit{{Without adversarial or external supervision}}} \\
ViTok-B/16$^{\star}$ & - & - & 1.63 \\
\addlinespace[3pt]
\textbf{\modelname{}-B (Ours)} & - & - & 1.01 \\
\quad + GAN decoder ft$^{\dagger}$ & \checkmark & - & \textbf{0.51} \\
\hdashline
\addlinespace[2pt]
w/ separate weights & - & - & 1.38 \\
\bottomrule
\end{tabular}
}
\par\vspace{1mm}
{\scriptsize $^{\star}$\,Stage-1 only (L2+LPIPS+KL). \; 
$^{\dagger}$\,Decoder-only ft, 16 epochs.}
\vspace{-5mm}
\end{table}

{With these choices in place, we find that both the weight-shared Generative Encoder and the separate encoder-denoiser variant exhibit high CKA/CKNNA alignment (Fig.~\ref{fig:cka_analysis_final}, left), indicating that tokenization and denoising are \textit{intrinsically aligned tasks} in our setting. This also clarifies the role of weight sharing: when the two tasks already align, parameter tying becomes a principled way to remove redundancy---especially in the reusable functional sublayers (attention and MLPs)---while retaining strong reconstruction and generation fidelity.}

{When analyzing the no-stop-gradient alternative (Fig.~\ref{fig:cka_analysis_final}, middle and right), we observe that, for both the weight-shared and separate encoder--denoiser settings, CKA and cosine-similarity alignment between the outputs of the tokenization and denoising pathways is reduced, relative to the stop-gradient variants, despite the denoising objective encouraging agreement at the final latent target. Further, Fig.~\ref{fig:denoising_analysis} shows that the no-stop-gradient models produce noticeably noisier intermediate denoised reconstructions. Taken together, these observations suggest that stopping denoising gradients through clean latent \textit{may} help preserve a more cleanly shared representation between tok. \& generation.}

\vspace{-2mm}
\paragraph{Entropy / Compression Analysis.}
We next study the encoder--denoiser relationship through the lens of \emph{compressibility}, motivated by a Minimum Description Length (MDL) perspective: if tokenization and denoising implement closely related computations, then a unified latent-generation program might admit a shorter description than two independently parameterized modules. Concretely, we estimate an empirical description-length proxy for model weights using per-tensor histogram entropy.

We begin with the separate encoder--denoiser setting. Compared to random weights, the total entropy of the encoder drops from $179.2$MB at random initialization to $121.9$MB after training, with both normalization parameters ($60.0 \rightarrow 30.7$MB) and functional attention/MLP parameters ($119.2 \rightarrow 91.2$MB) becoming substantially more structured as a result of training.

In the weight-shared Generative Encoder setting, the entropy of the functional attention/MLP parameters remains nearly \textit{unchanged} relative to the separate encoder ($91.2$MB $\rightarrow 90.8$MB), while the main increase is concentrated in normalization-related parameters, whose entropy rises modestly from $30.7$MB to $42.0$MB and closely matches that of the separate denoiser ($42.0$MB). Thus, \textit{unifying tokenization and denoising does not require a more complex functional backbone}; instead, the shared model reuses essentially the same attention/MLP computation and expresses the residual mode-specific adaptation primarily through normalization and scale parameters. This provides a complementary MDL-style interpretation of why sharing works in our setting: parameter tying may yield a shorter description of the joint latent-generation program, not by substantially altering the main reusable computation, but by preserving a common functional backbone while allocating only a small additional entropy budget to normalization. This interpretation aligns with our CKA analysis, since pathway alignment remains high while CKA is largely insensitive to norm/scale changes, suggesting that tokenization \& denoising differ more in feature calibration than in core representational geometry.
\vspace{2mm}

\section{Experimental Results}
\vspace{2mm}


{Can a single training job produce both a strong tokenizer and a strong generator?} 
In this section, we show that \modelname{} achieves near–state-of-the-art performance on both reconstruction and generation tasks across image and molecule modalities. 
 See Appendix for 
 more ablations.

\subsection{ImageNet-256 Results}
\vspace{1mm}

\begin{figure}[t]
\centering
\setlength{\tabcolsep}{0.5pt}
\renewcommand{\arraystretch}{0.5}
\begin{tabular}{ccccc}
\includegraphics[width=0.18\columnwidth]{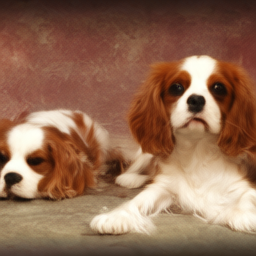} &
\includegraphics[width=0.18\columnwidth]{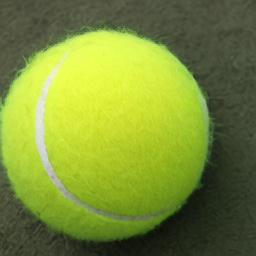} &
\includegraphics[width=0.18\columnwidth]{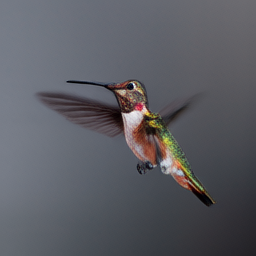} &
\includegraphics[width=0.18\columnwidth]{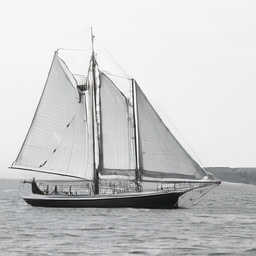} &
\includegraphics[width=0.18\columnwidth]{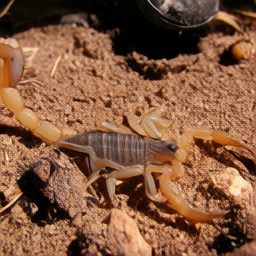} \\
\includegraphics[width=0.18\columnwidth]{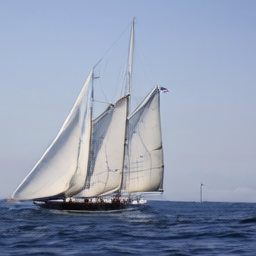} &
\includegraphics[width=0.18\columnwidth]{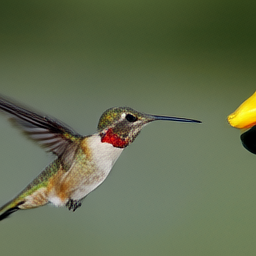} &
\includegraphics[width=0.18\columnwidth]{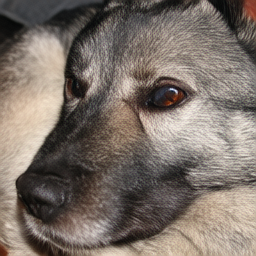} &
\includegraphics[width=0.18\columnwidth]{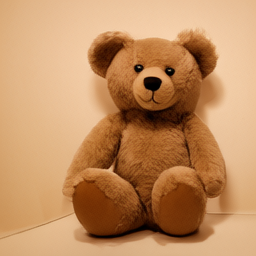} &
\includegraphics[width=0.18\columnwidth]{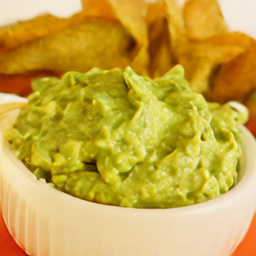} \\
\includegraphics[width=0.18\columnwidth]{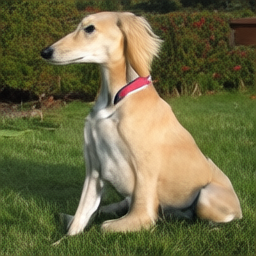} &
\includegraphics[width=0.18\columnwidth]{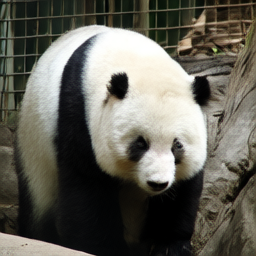} &
\includegraphics[width=0.18\columnwidth]{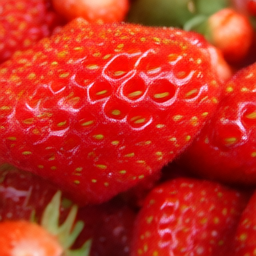} &
\includegraphics[width=0.18\columnwidth]{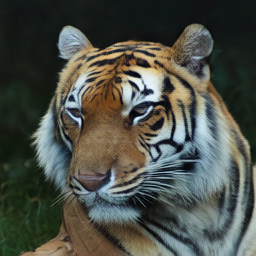} &
\includegraphics[width=0.18\columnwidth]{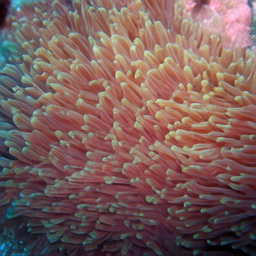} \\
\includegraphics[width=0.18\columnwidth]{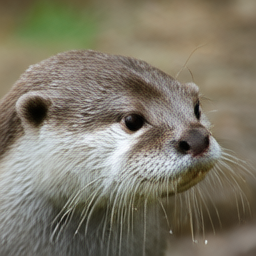} &
\includegraphics[width=0.18\columnwidth]{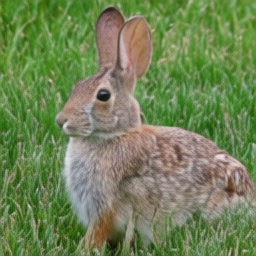} &
\includegraphics[width=0.18\columnwidth]{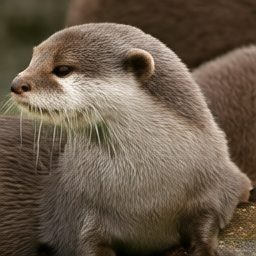} &
\includegraphics[width=0.18\columnwidth]{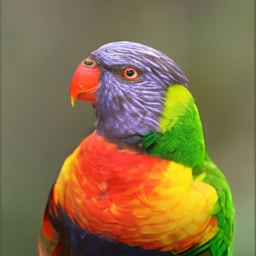} &
\includegraphics[width=0.18\columnwidth]{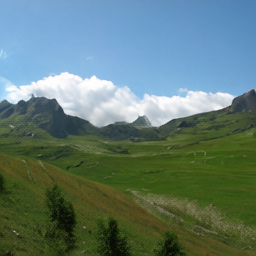} \\
\includegraphics[width=0.18\columnwidth]{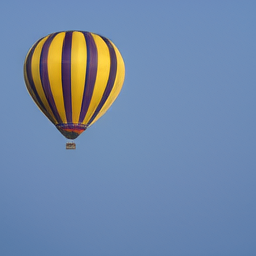} &
\includegraphics[width=0.18\columnwidth]{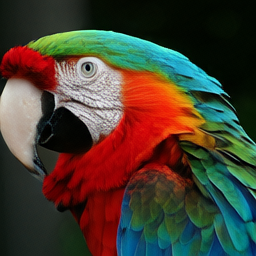} &
\includegraphics[width=0.18\columnwidth]{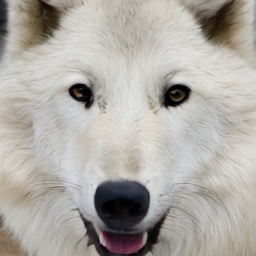} &
\includegraphics[width=0.18\columnwidth]{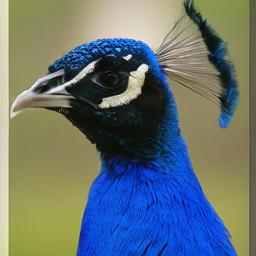} &
\includegraphics[width=0.18\columnwidth]{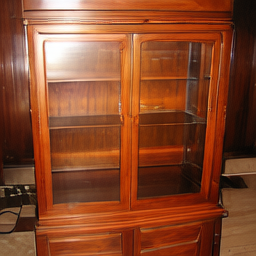} \\
\end{tabular}
\caption{\textbf{Selected samples from \modelname{}-XL.} Generated using 50 steps with CFG. This model achieves FID 1.75.
}
\label{fig:generation_samples}
\vspace{-4mm}
\end{figure}

\paragraph{Generation.} Tab.~\ref{tab:imagenet_results} summarizes our main generation results on ImageNet-256. The results indicate that truly end-to-end training of tokenization and generation is not only feasible, but also competitive. In particular, \modelname{}-B reaches an FID of 2.12, substantially improving over the single-stage baseline JiT-B/16~\cite{li2025jit} (FID 3.66). Increasing the model capacity further improves performance: \modelname{}-L (encoder size: L, default patch size: 16) reduces FID to \textbf{1.73}, surpassing two-stage approaches such as DiT-XL/2 (FID 2.27) and SiT-XL/2 (FID 2.06), suggesting that the unified setup continues to benefit from scale. Fig.~\ref{fig:generation_samples} shows representative samples from \modelname{}-XL (more uncurated class-conditional generations in Appendix). Unlike previous latent diffusion pipelines that train VAEs with GAN-based adversarial objectives, \modelname{} uses no adversarial loss.

Unlike RAE~\cite{rae2025diffusion} and REPA~\cite{yu2025repa}, which fundamentally rely on pretrained vision encoders, our single-stage approach reaches comparable performance while training from scratch, without requiring an external pretrained representation model.\footnote{Our ImageNet training uses LPIPS loss, which requires a pretrained VGG. However, (a) training VGG is inexpensive; (b) our molecule gen. results do not use LPIPS.} This simplicity—one encoder that serves both tokenization \& generation via weight sharing—makes the system {easier to train and deploy, 
reducing reliance on external pretrained components}. 

Compared with concurrent works including Latent Forcing \cite{baade2026latentforcingreorderingdiffusion} (LF-DiT-L as mentioned in Tab.~\ref{tab:imagenet_results}), Unified Latents \cite{heek2026unifiedlatentsultrain}, and Self-Flow \cite{CheferEsser2026selfflow}, our method is trained fully from scratch while achieving stronger generation FID. Unified Latents reports results only on ImageNet-512, with its best performance further relying on a second-stage diffusion fine-tuning step. Self-Flow, in contrast, builds on a pretrained DINO tokenizer and reports only unconditional generation results.
\vspace{2mm}

\vspace{-2mm}
\paragraph{Reconstruction.}
Tab.~\ref{tab:reconstruction} compares the reconstruction quality of \modelname{} against existing tokenizers. Most prior methods decouple reconstruction and generation, and low-rFID tokenizers such as VAEs and VQGANs typically rely on adversarial objectives in addition to reconstruction losses. More recent approaches further improve reconstruction fidelity by leveraging externally pretrained self-supervised encoders (e.g., DINOv2). As a reference point, a vanilla ViT autoencoder (ViTok-B/16 Stage~1~\cite{vitok2025}), trained from scratch with only L2+LPIPS+KL losses and no adversarial training, attains an rFID of 1.63.


Despite being trained jointly with a generative objective, \textbf{\modelname{}-B} (217M parameters) achieves an \textbf{rFID of 1.01} after 120 epochs, already outperforming the vanilla autoencoder baseline. A lightweight adversarial fine-tuning stage -- which freezes the Generative Encoder and updates only the decoder for 16 epochs---further reduces \textbf{rFID to 0.51}, surpassing all baselines, including RAE (0.58) and SD-VAE (0.62), without relying on any self-supervised pretraining. Finally, removing weight sharing (Tab.~\ref{tab:reconstruction} last row) substantially degrades reconstruction quality (rFID 1.38), further supporting the claim that shared parameterization benefits both reconstruction and generation.


\subsection{Beyond Vision: Application to Domains Without Pretrained Encoders}

Recent approaches such as REPA~\cite{yu2025repa} and RAE~\cite{rae2025diffusion} crucially depend on pretrained representation models, e.g., DINOv2 \cite{oquab2023dinov2}, to strengthen latent diffusion. This reliance makes their transfer to domains where such encoders are unavailable—or expensive to obtain—less straightforward, especially in settings with limited data or weaker pretraining ecosystems.

In contrast, our end-to-end formulation does not require pretrained encoders: tokenization and generation are learned jointly from scratch in a single training run
––making latent generative modeling applicable to domains where strong pretrained representation models do not exist.


We demonstrate this capability on QM9 molecule generation, a setting with no DINO-equivalent pretrained encoder. As shown in Tab.~\ref{tab:qm9_mol}, \modelname{} achieves state-of-the-art performance, matching or surpassing the All-atom Diffusion Transformer (ADiT) \cite{joshi2025allatom}—the current best method that relies on a separate VAE tokenizer. Notably, we obtain a 99.37\% reconstruction match rate (vs.\ 97.20\% for ADiT) and 99.71\% uniqueness among generated molecules (vs.\ 97.76\%), while training fully end-to-end and without any pretrained components.



\definecolor{myrule}{gray}{0.85}
\definecolor{headergray}{gray}{0.94}

\begin{table}[t]
    \centering
    \vspace{2mm}
    \caption{\textbf{QM9 molecule generation.} \modelname{}-S achieves the best reconstruction accuracy (99.37\% match) and uniqueness (99.71\%) under single-stage training. Crystal generation results on MP20 are provided in Appendix~\ref{app:mp20}.}
    \label{tab:qm9_mol}
    \scalebox{0.7}{
    \begin{tabular}{lcccc}
    \specialrule{.2em}{.1em}{.1em}
    \rowcolor{headergray}
    \multirow{2}{*}{} & \multicolumn{2}{c}{Reconstruction} & \multicolumn{2}{c}{Generation} \\
    \rowcolor{headergray}
     \multirow{-2}{*}{Method} & \makecell{Match\\(\%)} & \makecell{RMSD\\(\AA)} & \makecell{Valid\\(\%)} & \makecell{Unique\\(\%)} \\
    \specialrule{.1em}{.05em}{.05em} \vspace{-3mm}\\

    EDM \cite{hoogeboom2022equivariant} & -- & -- & 91.9 & 90.7 \\
    GeoLDM \cite{xu2023geometric} & -- & -- & 93.8 & 92.9 \\
    \hline \vspace{-3mm}\\

    ADiT Tokenizer \cite{joshi2025allatom} & 97.20 & 0.075 & -- & -- \\
    ADiT-S QM9-only \cite{joshi2025allatom} & -- & -- & \textbf{96.02} & 97.76 \\
    \hline \vspace{-3mm}\\

    \textbf{\modelname{}-S (Ours)} & \textbf{99.37} & \textbf{0.039} & 94.90 & \textbf{99.71} \\
    \specialrule{.1em}{.05em}{.05em}
    \end{tabular}
    }
    \vspace{-4mm}
\end{table}


These results further motivate studying true end-to-end training of tokenization and generation—where the two objectives are optimized jointly and gradients from each task shape the same representation space—as a means of enhancing latent diffusion models, rather than leveraging pretrained encoders trained on additional data.


\subsection{Training Efficiency}
\label{sec:efficiency}

We report total training FLOPs measured with gradient checkpointing enabled. For \modelname{}-B, each training sample costs approximately 3.5~TFLOPs (forward + backward), including one tokenization pass through \(\mathrm{GE}_\theta\) (512 tokens), fourteen denoising mini-batch passes (256 tokens each), one decoder pass, and one forward pass through a frozen VGG network for the LPIPS loss. Over 120 ImageNet epochs, \modelname{}-B requires approximately \(6.7\times 10^{20}\) FLOPs and reaches an FID of 2.18. Reducing the number of denoising iterations per reconstruction iteration can further lower training cost, at the expense of a modest increase in gFID. This is approx. \(\mathbf{15\times}\) cheaper than the end-to-end cost of methods that rely on pretrained DINOv2 encoders. RAE~\cite{rae2025diffusion} and LF-DiT~\cite{baade2026latentforcingreorderingdiffusion} both depend on DINOv2 features, whose ViT-g/14 pretraining and distillation together require approx. 27,000 A100-GPU-hours, corresponding to \(\sim 1.0\times 10^{22}\) model FLOPs \footnote{27,316 A100-GPU-hours \(\times\) 312~TFLOP/s (A100 BF16 peak) \(\times\) 0.4 model-FLOP utilization \(\approx 1.0\times 10^{22}\)~\cite{oquab2023dinov2}.}. This constitutes a fixed upfront cost inherited by any downstream method built on top of these features. In contrast, \modelname{} eliminates this overhead entirely by training from scratch.



Compared with standard two-stage latent diffusion models, our total compute is comparable: \modelname{}-B surpasses DiT-XL/2~\cite{peebles2023dit} (FID~2.27) at nearly matched total FLOPs (\(6.7\times 10^{20}\) vs.\ \(6.4\times 10^{20}\)), while using \(3\times\) fewer parameters (217M vs.\ 724M). In addition, \modelname{} jointly learns a tokenizer whose latent space is shaped by both reconstruction and generation objectives (Tab.~\ref{tab:reconstruction}).

Among single-stage methods, \modelname{}-B (\(6.7\times 10^{20}\) FLOPs, 217M parameters) achieves an FID of 2.18 at total compute comparable to JiT-G/16~\cite{li2025jit} (\(\sim 8.8\times 10^{20}\) FLOPs, 2B parameters, FID~1.82), while using approximately \(10\times\) fewer parameters. Moreover, \modelname{} produces a reusable latent tokenizer alongside the generator, a capability that pixel-space methods such as JiT do not offer.

\vspace{-2mm}
\section{Conclusion}
We present \modelname{}, a unified approach to joint tokenization and generation. Our encoder, termed the Generative Encoder, serves as both tokenizer and latent denoiser, with weights shared across the two objectives. This shared parameterization allows reconstruction and generation gradients to jointly shape the representation space, encouraging a common latent “language” that supports both tasks. \modelname{} is trained end-to-end in a single stage, with each iteration performing two forward passes through the same Generative Encoder: one for tokenization/reconstruction and one for latent denoising. Across ImageNet and molecule generation, \modelname{} achieves near-state-of-the-art fidelity: the base model reaches 2.12 gFID on ImageNet 256$\times$256, and scaling to XL improves this to 1.75 gFID. We further analyze the Generative Encoder through the lenses of representation alignment and compression.

More broadly, our results suggest two practical implications. First, it removes the reliance on pretrained encoders such as DINO for generative modeling, opening the door to latent generative modeling in domains where such encoders are unavailable. Second, our unified architecture is simpler and more efficient than conventional two-stage pipelines, reducing both implementation complexity and overall computational requirements.

\vspace{-1mm}
\section{Discussions}
\vspace{-1mm}
The core contribution of \modelname{} is to align tokenization and generation by training both over a shared latent space. The two objectives we consider are denoising and reconstruction. While reconstruction is a natural objective for learning compressed representations that preserve input information, \textit{exploring alternative objectives for tokenization beyond reconstruction} is an interesting direction for research---for example, jointly training the Generative Encoder with DINO- or JEPA-style objectives. This is especially appealing for robotics, where generative modeling can provide a useful world model of the environment. However, naively training such a world model on standard VAE latents may not yield \emph{actionable} latents that matter most for decision-making.

Another point worth discussing is the \textit{vision-language modeling capability of the Generative Encoder}. The Generative Encoder idea is loosely reminiscent of the classical wake-sleep algorithm, whose broader goal was to bridge discriminative and generative modeling. While \modelname{} achieves strong reconstruction and generation fidelity, the linear probing accuracy of the Generative Encoder remains comparable to that of other generative tokenizers, such as VAEs and VQGANs, at around $30\%$. We believe that linear probing (LP) alone may not be fully predictive of the discriminative strengths of highly compressed latent representations. In particular, stronger compression may require greater downstream decoding capacity before the representation becomes predictive for a given task. For this reason, evaluating the tokenizer in a VLM setting may provide a more informative picture of its discriminative capabilities than LP alone.

Furthermore, the results in Fig.~\ref{fig:denoising_analysis} suggest that weight sharing and end-to-end joint training of tokenization and generation may support further \textit{progress toward faster generative models}, potentially enabling high-quality one- to few-step generation. This also raises another interesting related question: can the process of mapping images to latents itself benefit from multiple iterative refinement loops? Prior work, such as ALIT~\cite{duggal2025adaptive}, has explored this direction and reported improvements in linear probing and token-level object binding with additional iterations.

\vspace{-2mm}
\section*{Acknowledgements}
\vspace{-2mm}
We are grateful to Jyo Pari, Shamit Lal, Tianyuan Zhang, Suwan Kim, Peter Holderrieth \& Qianwei Jia for fruitful discussions and constructive suggestions. We also thank Prof. Kaiming He for inspiring discussions on earlier iterations of this project.
This work is in part supported by MIT-IBM Watson AI Lab; ONR MURI grant \#033697-00007; the National Science Foundation under Cooperative Agreement PHY-2019786 (The NSF AI Institute for Artificial Intelligence and Fundamental Interactions, http://iaifi.org/). S.D. is further supported by Amazon AI Research Innovation Fellowship; X.B. is supported by MongoDB PhD fellowship. 

\vspace{-2mm}
\clearpage
\bibliography{arxiv}

@article{joshi2025allatom,
  title={All-atom Diffusion Transformers: Unified generative modelling of molecules and materials},
  author={Joshi, Chaitanya K. and Fu, Xiang and Liao, Yi-Lun and Gharakhanyan, Vahe and Miller, Benjamin Kurt and Sriram, Anuroop and Ulissi, Zachary W.},
  journal={arXiv preprint arXiv:2503.03965},
  year={2025}
}

@inproceedings{ho2020denoising,
  author = {Ho, Jonathan and Jain, Ajay and Abbeel, Pieter},
  booktitle = {Advances in Neural Information Processing Systems},
  pages = {6840--6851},
  publisher = {Curran Associates, Inc.},
  title = {Denoising Diffusion Probabilistic Models},
  volume = {33},
  year = {2020}
}

@article{kingma2014vae,
  title={Auto-Encoding Variational Bayes},
  author={Kingma, Diederik P. and Welling, Max},
  journal={arXiv preprint arXiv:1312.6114},
  year={2014}
}

@article{vandenoord2017vqvae,
  title={Neural Discrete Representation Learning},
  author={Van Den Oord, Aaron and Vinyals, Oriol and Kavukcuoglu, Koray},
  journal={Advances in Neural Information Processing Systems},
  volume={30},
  year={2017}
}

@InProceedings{esser2021vqgan,
  title={Taming Transformers for High-Resolution Image Synthesis},
  author={Esser, Patrick and Rombach, Robin and Ommer, Björn},
  booktitle={Proceedings of the IEEE/CVF Conference on Computer Vision and Pattern Recognition},
  pages={12873--12883},
  year={2021}
}

@article{he2022mae,
  title={Masked Autoencoders Are Scalable Vision Learners},
  author={He, Kaiming and Chen, Xinlei and Xie, Saining and Li, Yanghao and Doll{\'a}r, Piotr and Girshick, Ross},
  journal={Proceedings of the IEEE/CVF Conference on Computer Vision and Pattern Recognition},
  pages={16000--16009},
  year={2022}
}

@inproceedings{yu2025repa,
  title={Representation Alignment for Generation: Training Diffusion Transformers Is Easier Than You Think},
  author={Yu, Sihyun and Kwak, Sangkyung and Jang, Huiwon and Jeong, Jongheon and Huang, Jonathan and Shin, Jinwoo and Xie, Saining},
  booktitle={ICLR},
  year={2025}
}

@article{leng2025repae,
  title={REPA-E: Unlocking VAE for End-to-End Tuning with Latent Diffusion Transformers},
  author={Leng, Xingjian and Singh, Jaskirat and Hou, Yunzhong and Xing, Zhenchang and Xie, Saining and Zheng, Liang},
  journal={arXiv preprint arXiv:2504.10483},
  year={2025}
}

@article{rae2025diffusion,
  title={Diffusion Transformers with Representation Autoencoders},
  author={Zheng, Boyang and Ma, Nanye and Tong, Shengbang and Xie, Saining},
  journal={arXiv preprint arXiv:2510.11690},
  year={2025}
}

@article{tarflow2024,
  title={Normalizing Flows are Capable Generative Models},
  author={Zhai, Shuangfei and others},
  journal={arXiv preprint arXiv:2412.06329},
  year={2024}
}

@inproceedings{peebles2023dit,
  title={Scalable Diffusion Models with Transformers},
  author={Peebles, William and Xie, Saining},
  booktitle={Proceedings of the IEEE/CVF International Conference on Computer Vision (ICCV)},
  pages={4195--4205},
  year={2023},
  url={https://openaccess.thecvf.com/content/ICCV2023/html/Peebles_Scalable_Diffusion_Models_with_Transformers_ICCV_2023_paper.html}
}

@article{ma2024sit,
  title={SiT: Exploring Flow and Diffusion-based Generative Models with Scalable Interpolant Transformers},
  author={Ma, Nanye and Goldstein, Mark and Albergo, Michael S. and Boffi, Nicholas M. and Vanden-Eijnden, Eric and Xie, Saining},
  journal={arXiv preprint arXiv:2401.08740},
  year={2024},
  note={ECCV 2024},
  url={https://arxiv.org/abs/2401.08740}
}

@article{caron2021dino,
  title={Emerging Properties in Self-Supervised Vision Transformers},
  author={Caron, Mathilde and Touvron, Hugo and Misra, Ishan and J{\'e}gou, Herv{\'e} and Mairal, Julien and Bojanowski, Piotr and Joulin, Armand},
  journal={Proceedings of the IEEE/CVF International Conference on Computer Vision},
  pages={9650--9660},
  year={2021}
}

@article{oquab2023dinov2,
  title={DINOv2: Learning Robust Visual Features without Supervision},
  author={Oquab, Maxime and Darcet, Timoth{\'e}e and Moutakanni, Th{\'e}o and Vo, Huy and Szafraniec, Marc and Khalidov, Vasil and Fernandez, Pierre and Haziza, Daniel and Massa, Francisco and El-Nouby, Alaaeldin and Assran, Mahmoud and Ballas, Nicolas and Galuba, Wojciech and Howes, Russell and Huang, Po-Yao and Li, Shang-Wen and Misra, Ishan and Rabbat, Michael and Sharma, Vasu and Synnaeve, Gabriel and Xu, Hu and J{\'e}gou, Herv{\'e} and Mairal, Julien and Labatut, Patrick and Joulin, Armand and Bojanowski, Piotr},
  journal={arXiv preprint arXiv:2304.07193},
  year={2023},
  url={https://arxiv.org/abs/2304.07193}
}

@inproceedings{lipman2022flow,
  title={Flow Matching for Generative Modeling},
  author={Lipman, Yaron and Chen, Ricky T. Q. and Ben-Hamu, Heli and Nickel, Maximilian and Le, Matt},
  booktitle={International Conference on Learning Representations},
  year={2023},
  url={https://arxiv.org/abs/2210.02747}
}

@article{liu2023rectified,
  title={Flow Straight and Fast: Learning to Generate and Transfer Data with Rectified Flow},
  author={Liu, Xingchao and Gong, Chengyue and Liu, Qiang},
  journal={arXiv preprint arXiv:2209.03003},
  year={2023}
}

@article{li2025jit,
  title={Back to Basics: Let Denoising Generative Models Denoise},
  author={Li, Tianhong and He, Kaiming},
  journal={arXiv preprint arXiv:2511.13720},
  year={2025}
}

@inproceedings{hoogeboom2022equivariant,
  title={Equivariant Diffusion for Molecule Generation in 3D},
  author={Hoogeboom, Emiel and Satorras, V{\'\i}ctor Garcia and Vignac, Cl{\'e}ment and Welling, Max},
  booktitle={International Conference on Machine Learning},
  pages={8867--8887},
  year={2022},
  organization={PMLR}
}

@inproceedings{xu2023geometric,
  title={Geometric Latent Diffusion Models for 3D Molecule Generation},
  author={Xu, Minkai and Powers, Alexander S. and Dror, Ron O. and Ermon, Stefano and Leskovec, Jure},
  booktitle={International Conference on Machine Learning},
  pages={38592--38610},
  year={2023},
  organization={PMLR}
}

@article{ramakrishnan2014quantum,
  title={Quantum chemistry structures and properties of 134 kilo molecules},
  author={Ramakrishnan, Raghunathan and Dral, Pavlo O. and Rupp, Matthias and von Lilienfeld, O. Anatole},
  journal={Scientific Data},
  volume={1},
  number={1},
  pages={140022},
  year={2014},
  publisher={Nature Publishing Group}
}

@article{jain2013materials,
  title={Commentary: The Materials Project: A materials genome approach to accelerating materials innovation},
  author={Jain, Anubhav and Ong, Shyue Ping and Hautier, Geoffroy and Chen, Wei and Richards, William Davidson and Dacek, Stephen and Cholia, Shreyas and Gunter, Dan and Skinner, David and Ceder, Gerbrand and Persson, Kristin A.},
  journal={APL Materials},
  volume={1},
  number={1},
  pages={011002},
  year={2013},
  publisher={AIP Publishing}
}

@article{ong2013python,
  title={Python Materials Genomics (pymatgen): A robust, open-source python library for materials analysis},
  author={Ong, Shyue Ping and Richards, William Davidson and Jain, Anubhav and Hautier, Geoffroy and Kocher, Michael and Cholia, Shreyas and Gunter, Dan and Chevrier, Vincent L. and Persson, Kristin A. and Ceder, Gerbrand},
  journal={Computational Materials Science},
  volume={68},
  pages={314--319},
  year={2013},
  publisher={Elsevier}
}

@article{li2024autoregressive,
  title={Autoregressive Image Generation without Vector Quantization},
  author={Li, Tianhong and Li, He and Deng, Mingyang},
  journal={Advances in Neural Information Processing Systems},
  volume={37},
  year={2024}
}

@inproceedings{dosovitskiy2020image,
  title={An Image is Worth 16x16 Words: Transformers for Image Recognition at Scale},
  author={Dosovitskiy, Alexey and Beyer, Lucas and Kolesnikov, Alexander and Weissenborn, Dirk and Zhai, Xiaohua and Unterthiner, Thomas and Dehghani, Mostafa and Minderer, Matthias and Heigold, Georg and Gelly, Sylvain and Uszkoreit, Jakob and Houlsby, Neil},
  booktitle={International Conference on Learning Representations},
  year={2021},
  url={https://openreview.net/forum?id=YicbFdNTTy}
}

@article{chen2023pixart,
  title={PixArt-$\alpha$: Fast Training of Diffusion Transformer for Photorealistic Text-to-Image Synthesis},
  author={Chen, Junsong and Yu, Jincheng and Ge, Chongjian and Yao, Lewei and Xie, Enze and Wu, Yue and Wang, Zhongdao and Kwok, James and Luo, Ping and Lu, Huchuan and Li, Zhenguo},
  journal={arXiv preprint arXiv:2310.00426},
  year={2023},
  url={https://arxiv.org/abs/2310.00426}
}

@article{brown2020language,
  title={Language Models are Few-Shot Learners},
  author={Brown, Tom and Mann, Benjamin and Ryder, Nick and Subbiah, Melanie and Kaplan, Jared D. and Dhariwal, Prafulla and Neelakantan, Arvind and Shyam, Pranav and Sastry, Girish and Askell, Amanda and Agarwal, Sandhini and Herbert-Voss, Ariel and Krueger, Gretchen and Henighan, Tom and Child, Rewon and Ramesh, Aditya and Ziegler, Daniel and Wu, Jeffrey and Winter, Clemens and Hesse, Chris and Chen, Mark and Sigler, Eric and Litwin, Mateusz and Gray, Scott and Chess, Benjamin and Clark, Jack and Berner, Christopher and McCandlish, Sam and Radford, Alec and Sutskever, Ilya and Amodei, Dario},
  journal={Advances in Neural Information Processing Systems},
  volume={33},
  pages={1877--1901},
  year={2020},
  url={https://papers.nips.cc/paper/2020/hash/1457c0d6bfcb4967418bfb8ac142f64a-Abstract.html}
}

@article{koroteev2021bert,
  title={BERT: a review of applications in natural language processing and understanding},
  author={Koroteev, Mikhail V},
  journal={arXiv preprint arXiv:2103.11943},
  year={2021}
}

@inproceedings{radford2021learning,
  title={Learning Transferable Visual Models From Natural Language Supervision},
  author={Radford, Alec and Kim, Jong Wook and Hallacy, Chris and Ramesh, Aditya and Goh, Gabriel and Agarwal, Sandhini and Sastry, Girish and Askell, Amanda and Mishkin, Pamela and Clark, Jack and Krueger, Gretchen and Sutskever, Ilya},
  booktitle={Proceedings of the 38th International Conference on Machine Learning},
  series={Proceedings of Machine Learning Research},
  volume={139},
  pages={8748--8763},
  year={2021},
  publisher={PMLR},
  url={https://proceedings.mlr.press/v139/radford21a.html}
}

@inproceedings{esser2024scaling,
  title={Scaling Rectified Flow Transformers for High-Resolution Image Synthesis},
  author={Esser, Patrick and Kulal, Sumith and Blattmann, Andreas and Entezari, Rahim and M{\"u}ller, Jonas and Saini, Harry and Levi, Yam and Lorenz, Dominik and Sauer, Axel and Boesel, Frederic and Podell, Dustin and Dockhorn, Tim and English, Zion and Rombach, Robin},
  booktitle={Proceedings of the 41st International Conference on Machine Learning},
  series={Proceedings of Machine Learning Research},
  volume={235},
  pages={12606--12633},
  year={2024},
  publisher={PMLR},
  url={https://proceedings.mlr.press/v235/esser24a.html}
}

@article{polyak2024movie,
  title={Movie gen: A cast of media foundation models},
  author={Polyak, Adam and Zohar, Amit and Brown, Andrew and Tjandra, Andros and Sinha, Animesh and Lee, Ann and Vyas, Apoorv and Shi, Bowen and Ma, Chih-Yao and Chuang, Ching-Yao and others},
  journal={arXiv preprint arXiv:2410.13720},
  year={2024}
}

@article{wan2025wan,
  title={Wan: Open and advanced large-scale video generative models},
  author={Wan, Team and Wang, Ang and Ai, Baole and Wen, Bin and Mao, Chaojie and Xie, Chen-Wei and Chen, Di and Yu, Feiwu and Zhao, Haiming and Yang, Jianxiao and others},
  journal={arXiv preprint arXiv:2503.20314},
  year={2025}
}

@inproceedings{song2020score,
  title={Score-Based Generative Modeling through Stochastic Differential Equations},
  author={Song, Yang and Sohl-Dickstein, Jascha and Kingma, Diederik P and Kumar, Abhishek and Ermon, Stefano and Poole, Ben},
  booktitle={International Conference on Learning Representations},
  year={2021},
  url={https://openreview.net/forum?id=PxTIG12RRHS}
}

@inproceedings{hoogeboom2023simple,
  title={simple diffusion: End-to-end diffusion for high resolution images},
  author={Hoogeboom, Emiel and Heek, Jonathan and Salimans, Tim},
  booktitle={International Conference on Machine Learning},
  pages={13213--13232},
  year={2023},
  organization={PMLR}
}

@article{chen2023importance,
  title={On the importance of noise scheduling for diffusion models},
  author={Chen, Ting},
  journal={arXiv preprint arXiv:2301.10972},
  year={2023}
}

@article{kingma2023understanding,
  title={Understanding diffusion objectives as the elbo with simple data augmentation},
  author={Kingma, Diederik and Gao, Ruiqi},
  journal={Advances in Neural Information Processing Systems},
  volume={36},
  pages={65484--65516},
  year={2023}
}

@inproceedings{kornblith2019similarity,
  title={Similarity of Neural Network Representations Revisited},
  author={Kornblith, Simon and Norouzi, Mohammad and Lee, Honglak and Hinton, Geoffrey},
  booktitle={Proceedings of the 36th International Conference on Machine Learning},
  series={Proceedings of Machine Learning Research},
  volume={97},
  pages={3519--3529},
  year={2019},
  publisher={PMLR},
  url={https://proceedings.mlr.press/v97/kornblith19a.html}
}

@article{goodfellow2020generative,
  title={Generative adversarial networks},
  author={Goodfellow, Ian and Pouget-Abadie, Jean and Mirza, Mehdi and Xu, Bing and Warde-Farley, David and Ozair, Sherjil and Courville, Aaron and Bengio, Yoshua},
  journal={Communications of the ACM},
  volume={63},
  number={11},
  pages={139--144},
  year={2020},
  publisher={ACM New York, NY, USA}
}

@article{hoogeboom2024simpler,
  title={Simpler diffusion (sid2): 1.5 fid on imagenet512 with pixel-space diffusion},
  author={Hoogeboom, Emiel and Mensink, Thomas and Heek, Jonathan and Lamerigts, Kay and Gao, Ruiqi and Salimans, Tim},
  journal={arXiv preprint arXiv:2410.19324},
  year={2024}
}

@article{chen2025pixelflow,
  title={PixelFlow: Pixel-Space Generative Models with Flow},
  author={Chen, Shoufa and Ge, Chongjian and Zhang, Shilong and Sun, Peize and Luo, Ping},
  journal={arXiv preprint arXiv:2504.07963},
  year={2025}
}

@article{wang2025pixnerd,
  title={Pixnerd: Pixel neural field diffusion},
  author={Wang, Shuai and Gao, Ziteng and Zhu, Chenhui and Huang, Weilin and Wang, Limin},
  journal={arXiv preprint arXiv:2507.23268},
  year={2025}
}

@article{ba2016layer,
  title={Layer normalization},
  author={Ba, Jimmy Lei and Kiros, Jamie Ryan and Hinton, Geoffrey E},
  journal={arXiv preprint arXiv:1607.06450},
  year={2016}
}

@article{yu2025pixeldit,
  title={PixelDiT: Pixel Diffusion Transformers for Image Generation},
  author={Yu, Yongsheng and Xiong, Wei and Nie, Weili and Sheng, Yichen and Liu, Shiqiu and Luo, Jiebo},
  journal={arXiv preprint arXiv:2511.20645},
  year={2025}
}

@article{chen2025dip,
  title={DiP: Taming Diffusion Models in Pixel Space},
  author={Chen, Zhennan and Zhu, Junwei and Chen, Xu and Zhang, Jiangning and Hu, Xiaobin and Zhao, Hanzhen and Wang, Chengjie and Yang, Jian and Tai, Ying},
  journal={arXiv preprint arXiv:2511.18822},
  year={2025}
}

@misc{heek2026unifiedlatentsultrain,
      title={Unified Latents (UL): How to train your latents}, 
      author={Jonathan Heek and Emiel Hoogeboom and Thomas Mensink and Tim Salimans},
      year={2026},
      eprint={2602.17270},
      archivePrefix={arXiv},
      primaryClass={cs.LG},
      url={https://arxiv.org/abs/2602.17270}, 
}

@misc{baade2026latentforcingreorderingdiffusion,
      title={Latent Forcing: Reordering the Diffusion Trajectory for Pixel-Space Image Generation}, 
      author={Alan Baade and Eric Ryan Chan and Kyle Sargent and Changan Chen and Justin Johnson and Ehsan Adeli and Li Fei-Fei},
      year={2026},
      eprint={2602.11401},
      archivePrefix={arXiv},
      primaryClass={cs.CV},
      url={https://arxiv.org/abs/2602.11401}, 
}

@article{CheferEsser2026selfflow,
      title={Self-Supervised Flow Matching for Scalable Multi-Modal Synthesis},
      author={Hila Chefer and Patrick Esser and Dominik Lorenz and Dustin Podell and Vikash Raja and Vinh Tong and Antonio Torralba and Robin Rombach},
      journal = {arXiv preprint arXiv:2603.06507},
      year={2026},
}

@article{vitok2025,
      title={Learnings from Scaling Visual Tokenizers for Reconstruction and Generation},
      author={Philippe Hansen-Estruch and David Yan and Ching-Yao Chung and Orr Zohar and Jialiang Wang and Sriram Vishwanath and Peter Vajda and Xinlei Chen},
      journal={arXiv preprint arXiv:2501.09755},
      year={2025},
}

@inproceedings{chen2018neural,
  title={Neural Ordinary Differential Equations},
  author={Chen, Ricky T. Q. and Rubanova, Yulia and Bettencourt, Jesse and Duvenaud, David},
  booktitle={Advances in Neural Information Processing Systems},
  volume={31},
  year={2018},
  url={https://papers.nips.cc/paper/2018/hash/69386f6bb1dfed68692a24c8686939b9-Abstract.html}
}

@inproceedings{tian2024visual,
  title={Visual Autoregressive Modeling: Scalable Image Generation via Next-Scale Prediction},
  author={Tian, Keyu and Jiang, Yi and Yuan, Zehuan and Peng, Bingyue and Wang, Liwei},
  booktitle={Advances in Neural Information Processing Systems},
  volume={37},
  year={2024},
  url={https://arxiv.org/abs/2404.02905}
}

@inproceedings{
duggal2025adaptive,
title={Adaptive Length Image Tokenization via Recurrent Allocation},
author={Shivam Duggal and Phillip Isola and Antonio Torralba and William T. Freeman},
booktitle={The Thirteenth International Conference on Learning Representations},
year={2025},
url={https://openreview.net/forum?id=mb2ryuZ3wz}
}
\bibliographystyle{icml2026}

\newpage
\onecolumn
\appendix
\section*{Appendix}


In this appendix, we first provide more details on the reconstruction fidelity results in Sec.~\ref{app:reconstruction}. Next, we share evaluation details (Sec.~\ref{app:evaluation}), along with additional uncurated samples generated by our model \modelname{}-XL shown in Fig.~\ref{fig:app_uncurated_samples}. We also provide architectural and training details in Tab.~\ref{tab:app_architecture_config} and Tab.~\ref{tab:app_training_config}. Finally, Sec.~\ref{app:molecule_generation} presents additional results on the molecule generation task and ablations on ImageNet.




\section{Reconstruction Fidelity Details}
\label{app:reconstruction}

Table~\ref{tab:reconstruction} in the main paper summarizes our reconstruction results. Here, we provide additional details on the adversarial fine-tuning procedure that reduces rFID from 1.01 to 0.51, without changing gFID.

\vspace{-3mm}
\paragraph{GAN Decoder Fine-Tuning.}
After \modelname{} joint training converges, we optionally apply a lightweight adversarial fine-tuning stage that targets \emph{only} the decoder. Concretely, we freeze the Generative Encoder entirely and train the decoder with an additional GAN loss for 16 epochs. The discriminator is initialized from our Generative Encoder, which already encodes rich semantic features from joint training; this eliminates the need for an external pretrained network (e.g., DINOv2) as the discriminator backbone. Due to limited compute budget, we did not explore fine-tuning both the encoder and decoder jointly with adversarial training.


\vspace{-3mm}
\paragraph{Effect of Weight Sharing on Reconstruction.}
As shown in Table~\ref{tab:reconstruction}, removing weight sharing between the encoder and denoiser (in the stop-gradient setting) degrades rFID from 1.01 to 1.38. We attribute this to the fact that in the shared-weight setting, the generation objective acts as an implicit regularizer on the encoder, encouraging latent representations that are both reconstructive and generatively useful. Separate weights remove this coupling, leading to a less structured latent space. That said, the separate encoder–denoiser variant without stop-gradient achieved a much lower rFID, indicating that joint training of tokenization and generation is beneficial.




\section{Evaluation Protocol}
\label{app:evaluation}

\begin{figure*}[h!]
\centering
\setlength{\tabcolsep}{3pt}
\renewcommand{\arraystretch}{0.8}
\begin{tabular}{cc}
\includegraphics[width=0.48\textwidth]{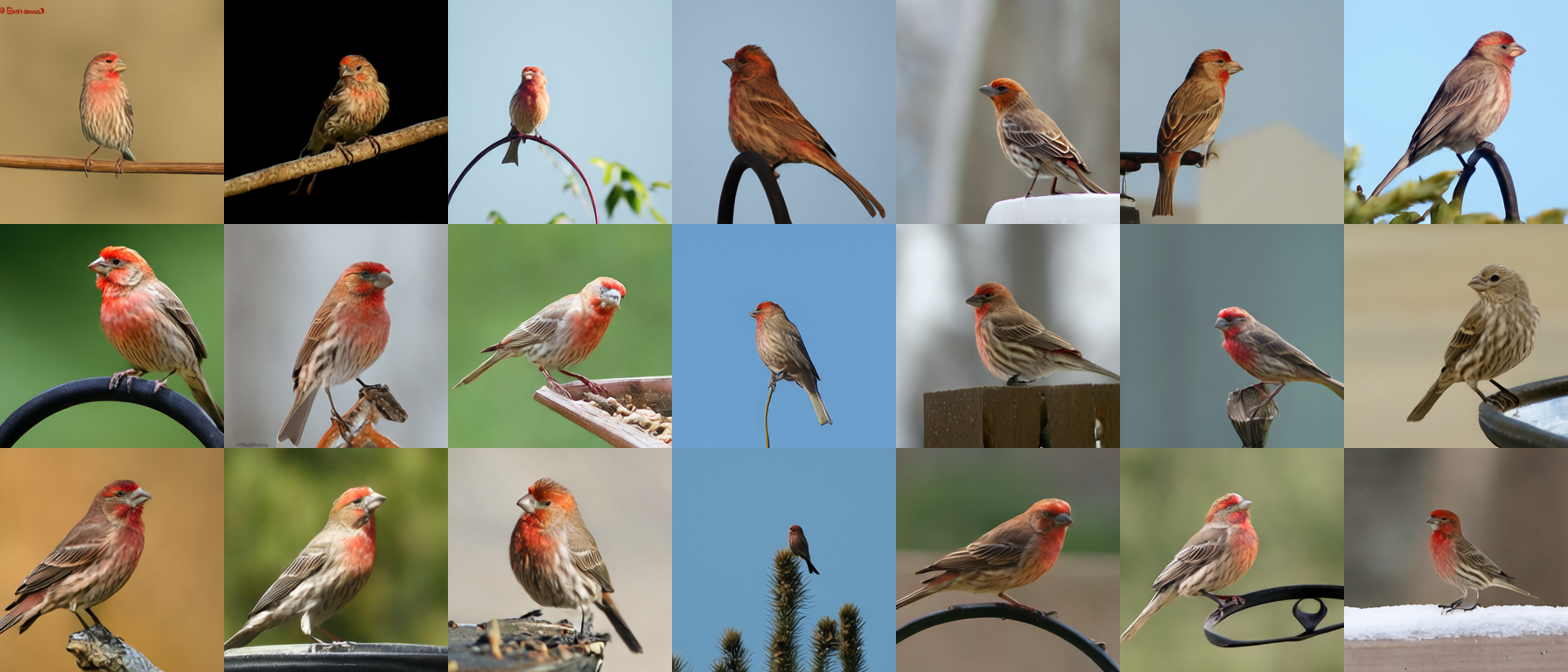} &
\includegraphics[width=0.48\textwidth]{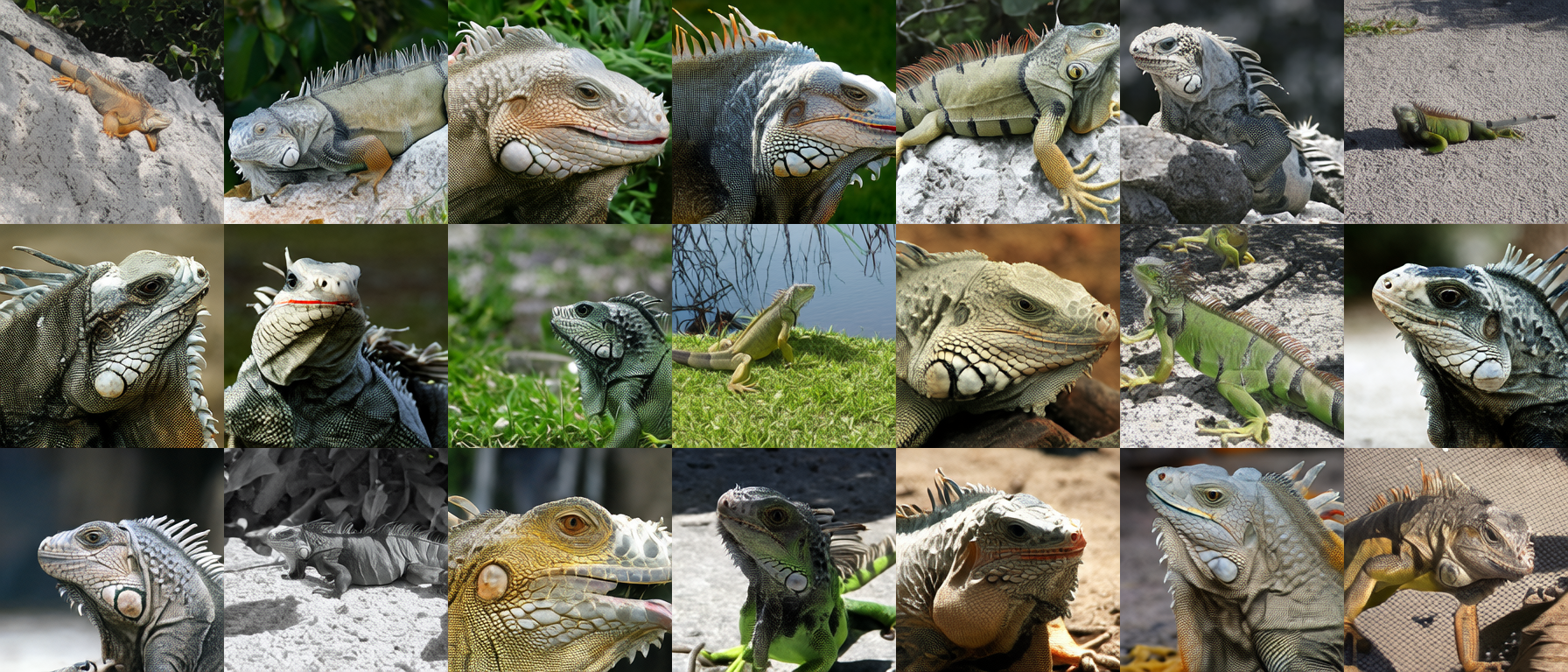} \\
\small class 12: house finch, linnet, Carpodacus mexicanus & \small class 39: common iguana, iguana, Iguana iguana \\[3pt]
\includegraphics[width=0.48\textwidth]{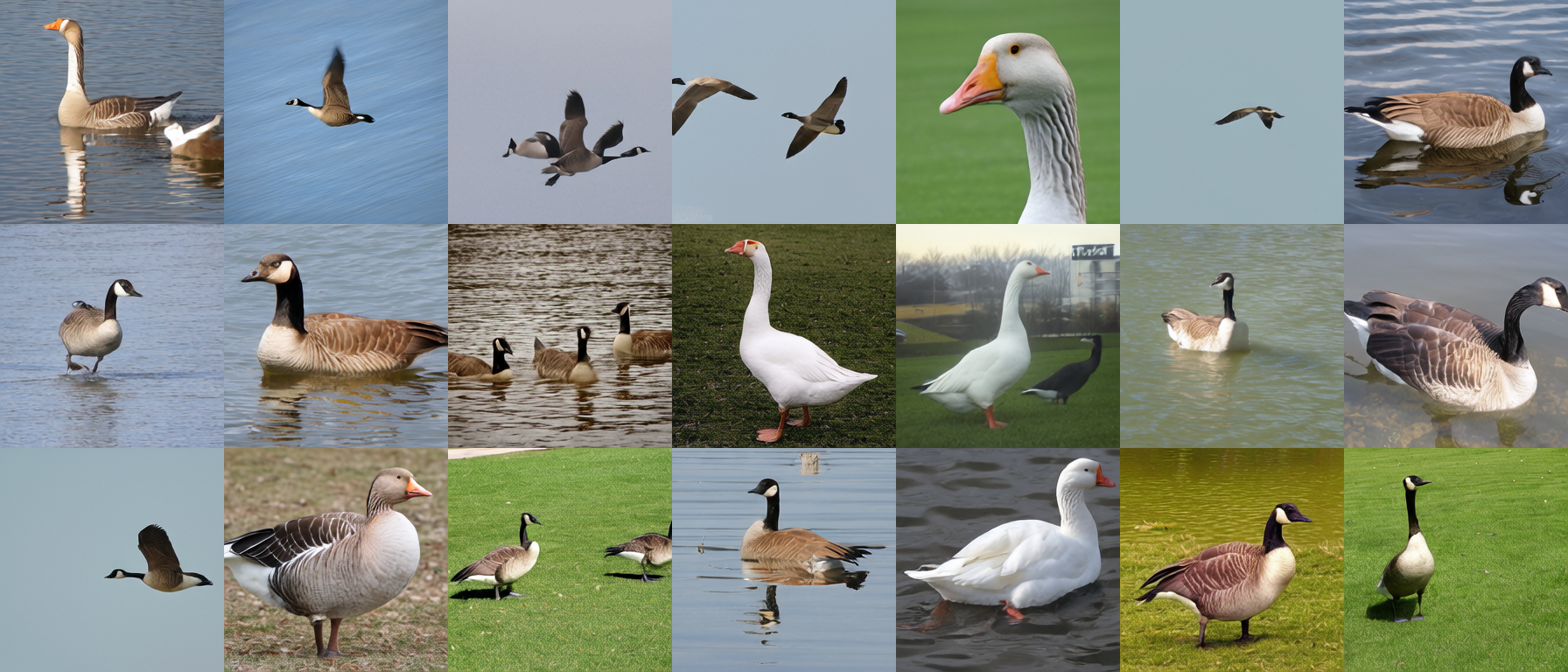} &
\includegraphics[width=0.48\textwidth]{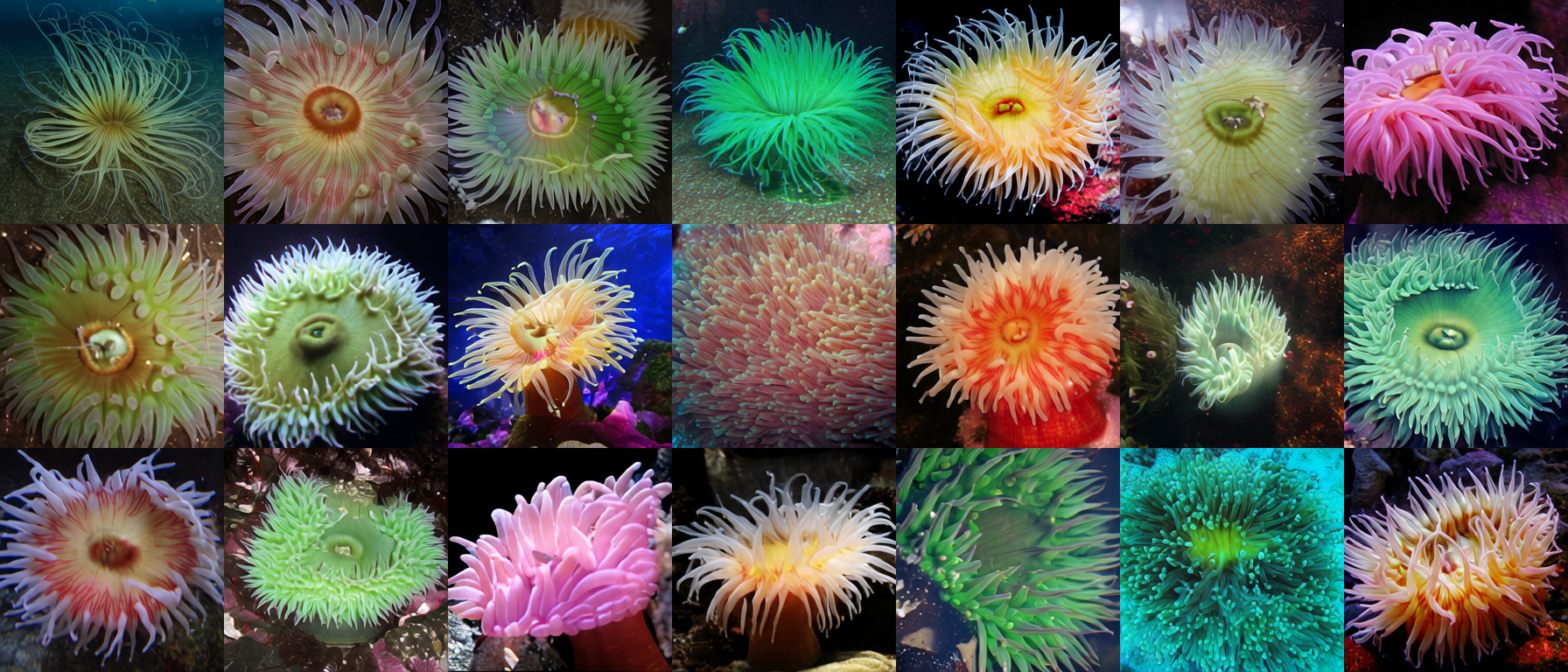} \\
\small class 99: goose & \small class 108: sea anemone, anemone \\[3pt]
\includegraphics[width=0.48\textwidth]{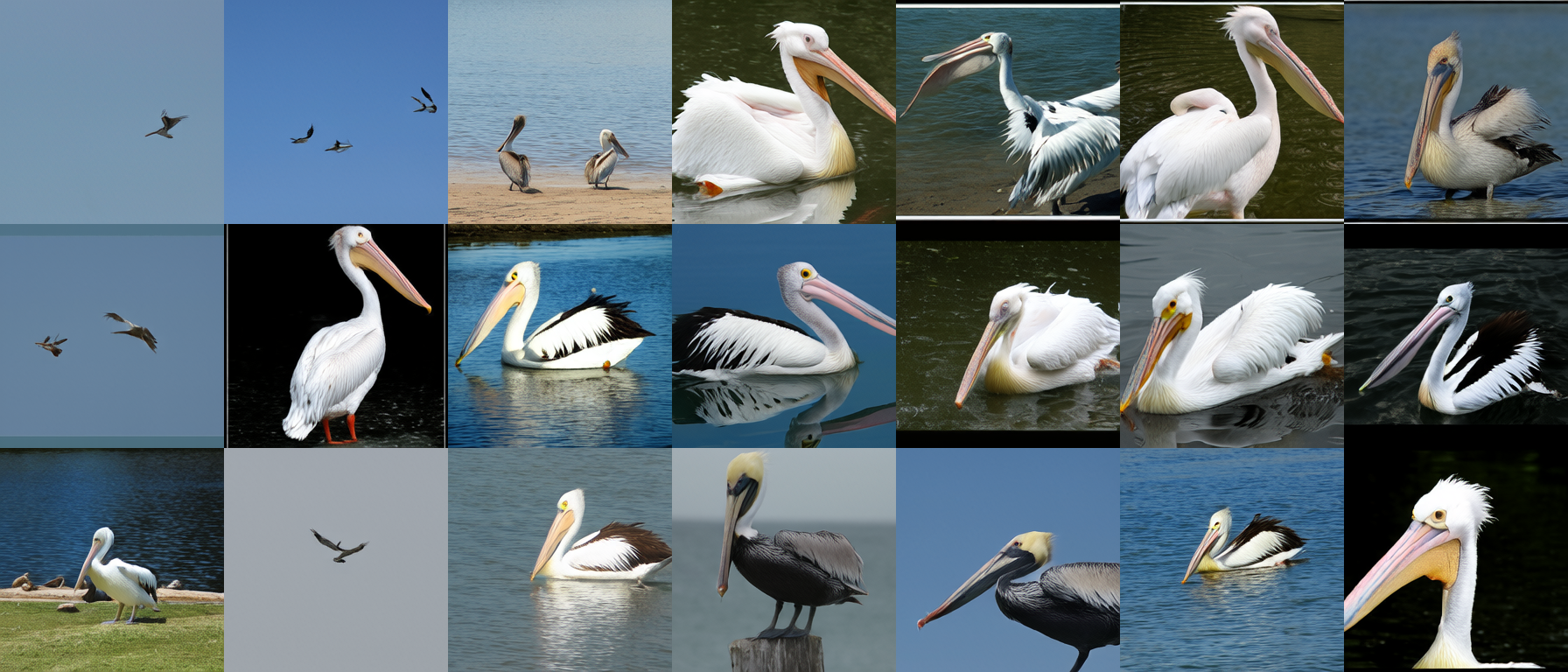} &
\includegraphics[width=0.48\textwidth]{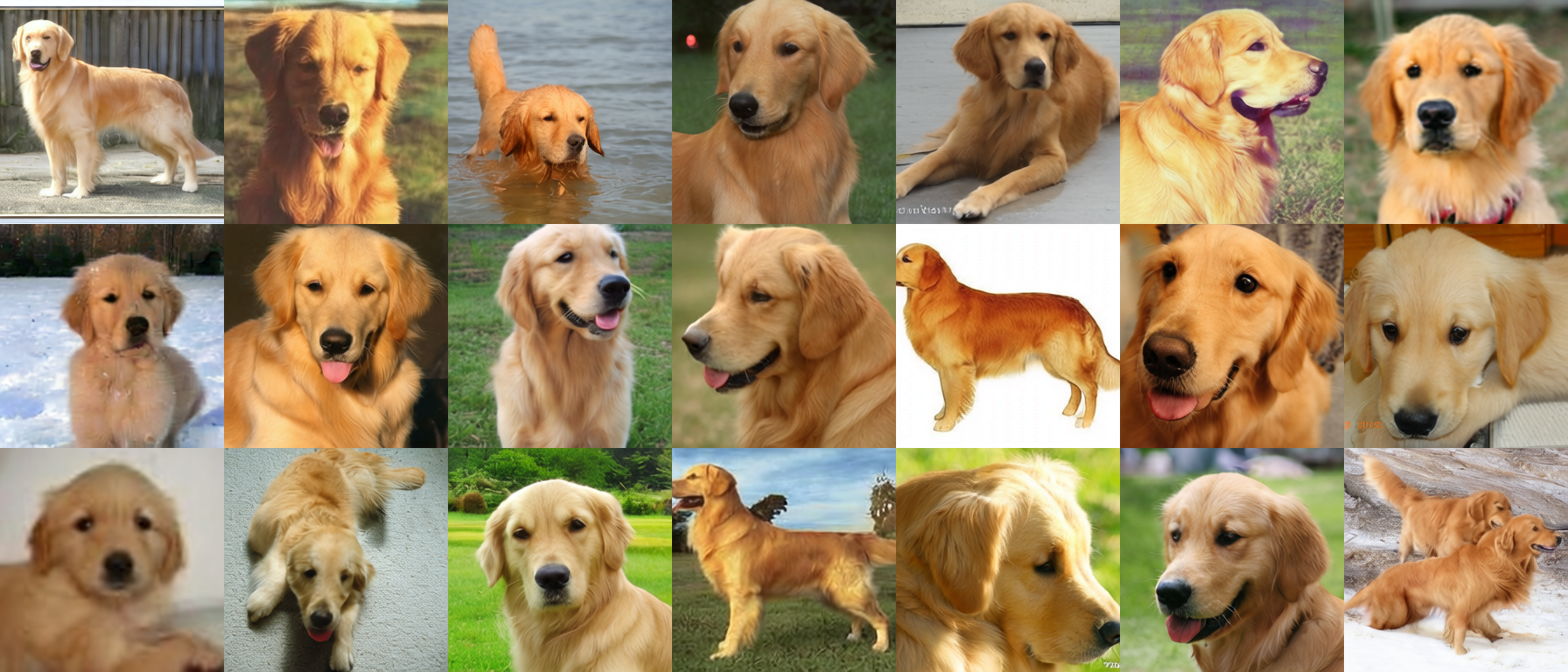} \\
\small class 144: pelican & \small class 207: golden retriever \\[3pt]
\includegraphics[width=0.48\textwidth]{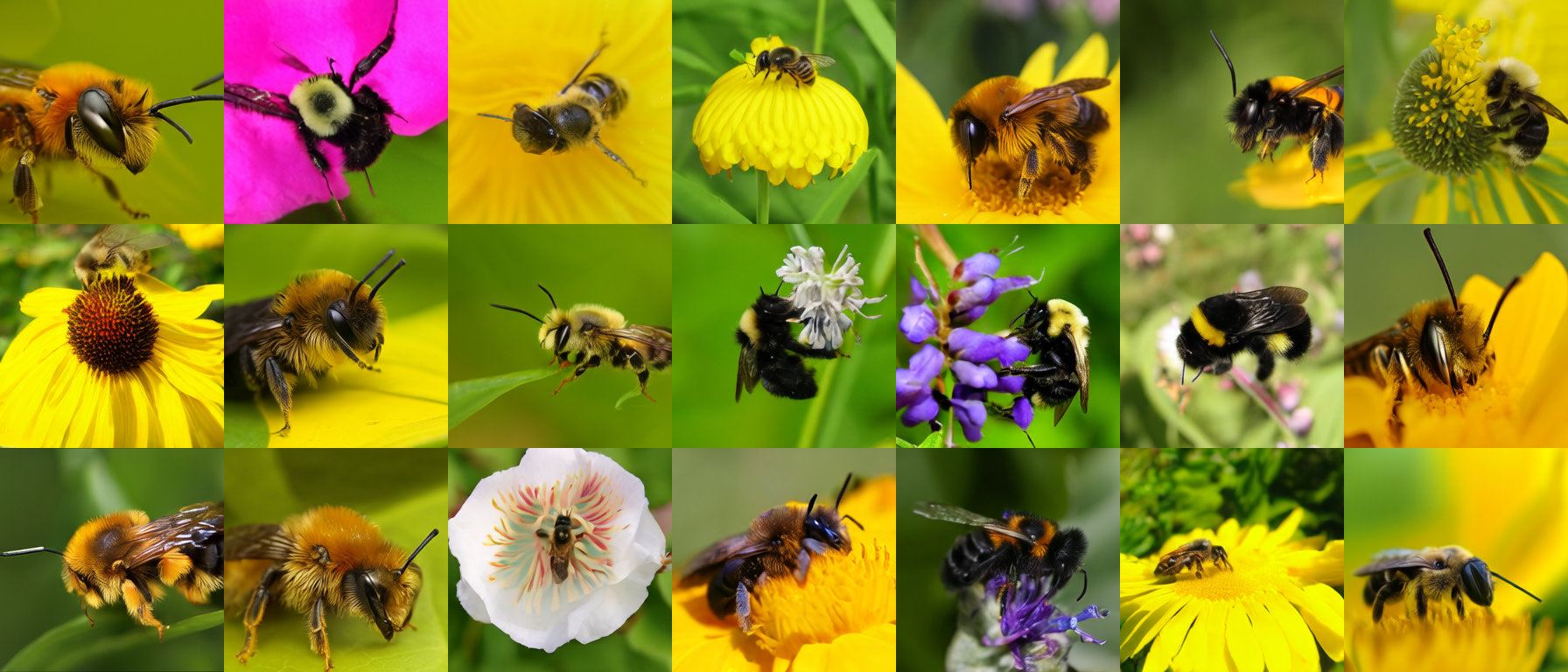} &
\includegraphics[width=0.48\textwidth]{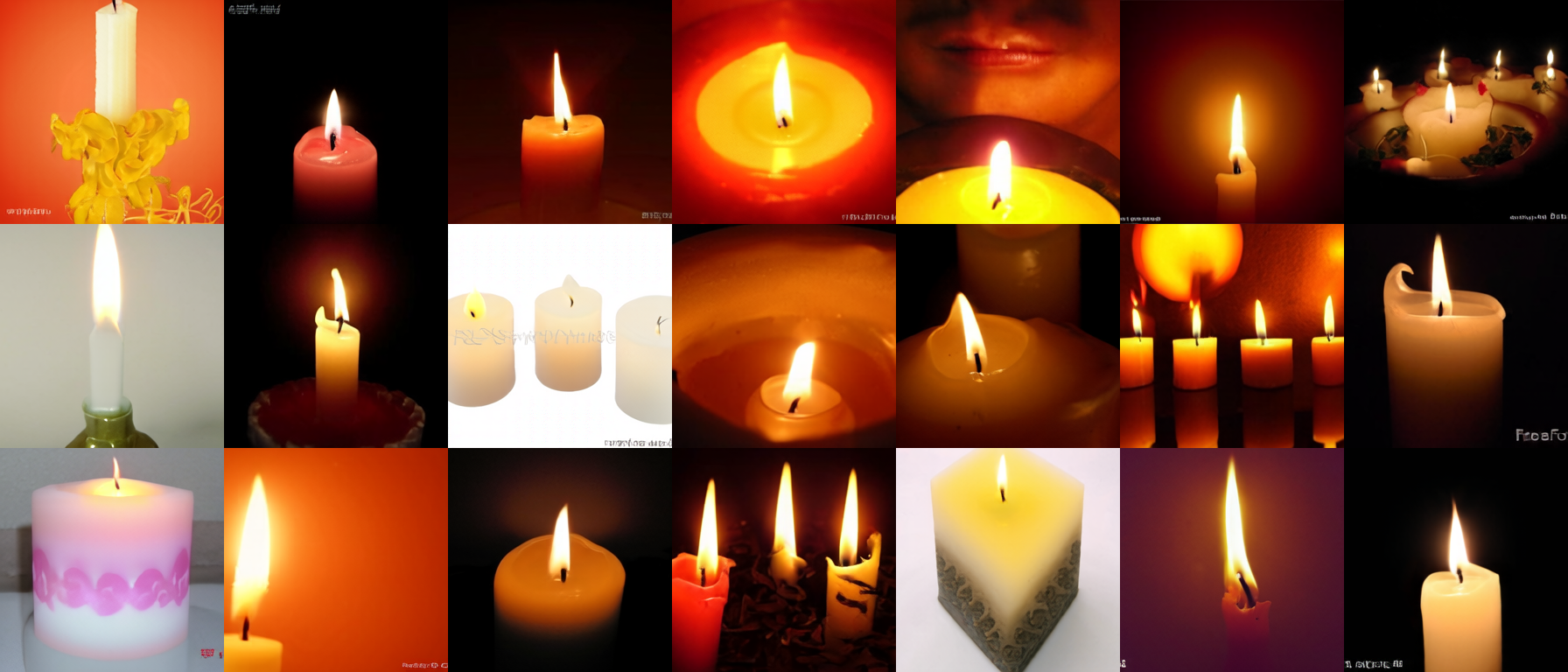} \\
\small class 309: bee & \small class 470: candle, taper, wax light \\[3pt]
\includegraphics[width=0.48\textwidth]{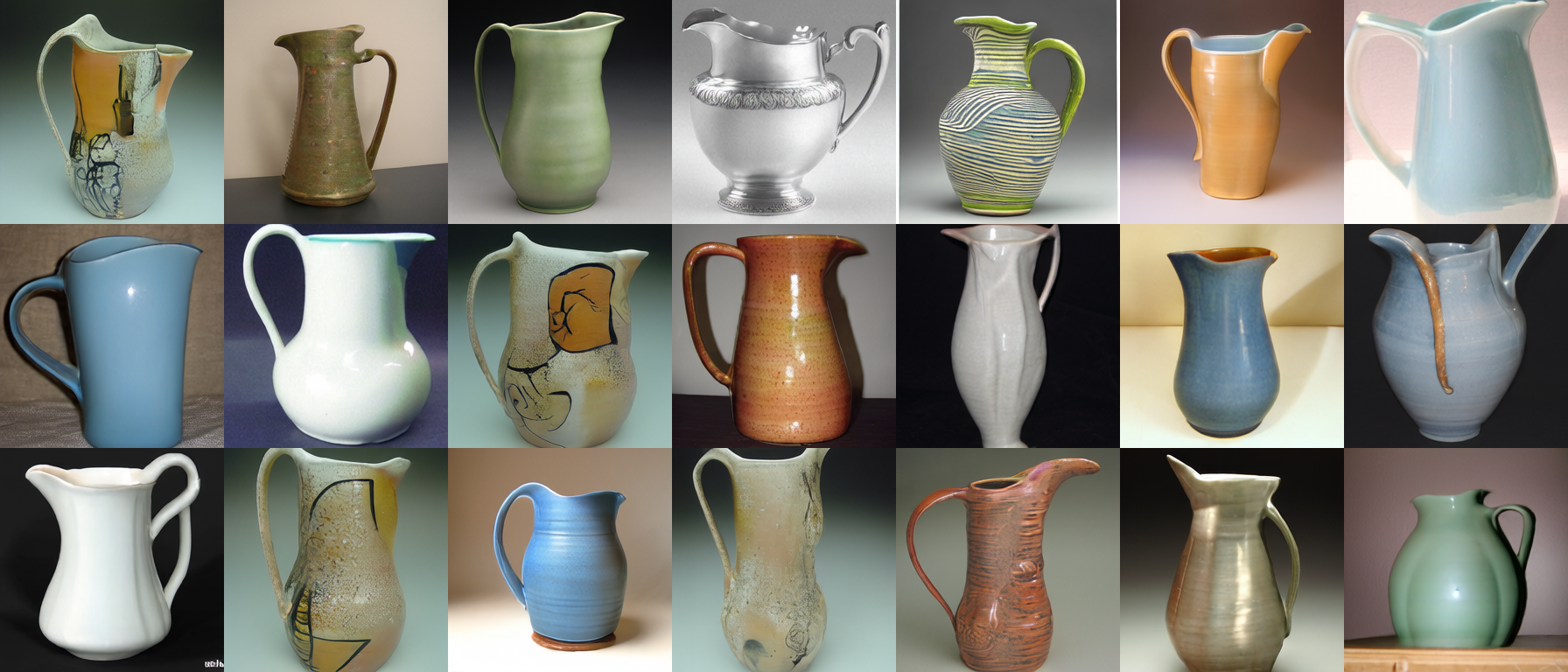} &
\includegraphics[width=0.48\textwidth]{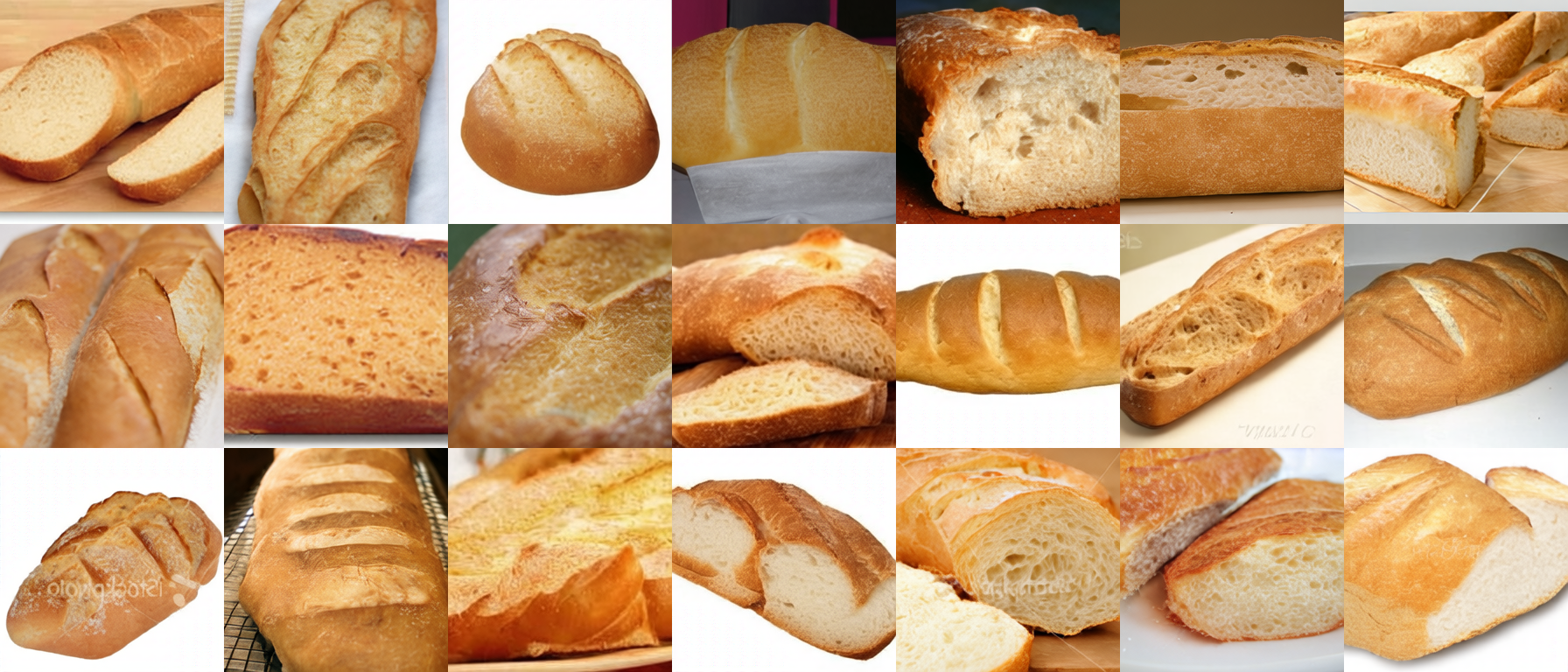} \\
\small class 725: pitcher, ewer & \small class 930: French loaf\\
\end{tabular}
\caption{\textbf{Uncurated, class-conditional samples on ImageNet 256×256 using \modelname{}-XL.}
We show images using CFG $\omega=4.0$.
Each grid contains 21 randomly sampled images, demonstrating consistent quality across diverse categories including animals, objects, and scenes.}
\label{fig:app_uncurated_samples}
\end{figure*}

For reproducibility, we detail the full evaluation protocol used for all ImageNet-256 generation results. See also Fig.~\ref{fig:app_uncurated_samples} for additional uncurated samples generated by \modelname{}-XL.

\paragraph{FID Computation.}
We compute Fr\'{e}chet Inception Distance (FID) using the \texttt{torch-fidelity} library with InceptionV3 features. Reference statistics are computed on the full ImageNet-1K training set (1281167 images). All reported FID scores use 50K generated samples.

\paragraph{Sampling Protocol.}
We adopt class-balanced sampling: exactly 50 images are generated per class for the 1K ImageNet classes, totaling 50K images. This follows the protocol used by VAR~\cite{tian2024visual}, MAR~\cite{li2024autoregressive}, and RAE~\cite{rae2025diffusion}, among others. As shown in RAE (Table~14), class-balanced sampling yields about 0.1 lower FID than the uniform random class sampling used in some prior work (e.g., DiT~\cite{peebles2023dit}, SiT~\cite{ma2024sit}). We note this systematic difference when comparing absolute FID values across methods.

\paragraph{ODE Solver and Inference Details.}
At inference time, we solve the probability flow ODE over the interval $[0.1,\, 1.0]$ with classifier-free guidance. We sweep the CFG scale $\omega$ from 1.0 to 4.0 in increments of 0.2 and report best FID for each model.

For all reported FID numbers in the main paper (Tab.~\ref{tab:imagenet_results}), we use the \textbf{adaptive fifth-order Dormand--Prince solver} (\texttt{dopri5}, from \texttt{torchdiffeq}~\cite{chen2018neural}), following the default configuration of the SiT codebase~\cite{ma2024sit}. For our model, we observe that \texttt{dopri5} uses ${\sim}$108 NFEs on average per sample (estimated from wall-clock timing), compared to exactly 100 NFEs for the fixed-step Heun solver with 50 steps.

Tab.~\ref{tab:app_solver_comparison} compares FID under different evaluation protocols for the same \modelname{}-B checkpoint. Switching to a fixed-step second-order Heun solver with 50 steps (100 NFEs) yields FID within ${\sim}0.05$ of \texttt{dopri5}, consistent with prior observations that flow-matching models produce near-linear trajectories that are well approximated by low-order fixed-step integrators~\cite{lipman2022flow, liu2023rectified}. This small gap is also consistent with the SiT authors' report that the FID difference between \texttt{dopri5} and fixed-step solvers is ${<}0.1$.\footnote{See \url{https://github.com/willisma/SiT/issues/21}.}

\begin{table}[h]
\centering
\caption{\textbf{Effect of evaluation protocol on reported FID.} All rows use the same \modelname{}-B checkpoint (240 epochs). ``Balanced'' denotes 50 images per class; ``Random'' denotes uniformly sampled class labels. NFE = number of function evaluations.}
\label{tab:app_solver_comparison}
\small
\begin{tabular}{llccc}
\toprule
\textbf{ODE Solver} & \textbf{Class Sampling} & \textbf{NFE} & \textbf{FID} $\downarrow$ & \textbf{IS} $\uparrow$ \\
\midrule
Heun (50 steps) & Balanced & 100 & 2.789 & 287.6 \\
dopri5 (adaptive) & Balanced & $\sim$108 & 2.735 & 268.1 \\
dopri5 (adaptive) & Random & $\sim$108 & 2.885 & 274.7 \\
\bottomrule
\end{tabular}
\end{table}










\section{Additional Results}
\label{app:molecule_generation}

\subsection{QM9 Molecular Generation}
\label{app:qm9_setup}

The QM9 dataset~\citep{ramakrishnan2014quantum} contains approximately 130K stable small organic molecules with up to 9 heavy atoms from the set \{C, N, O, F\}. Following \citet{joshi2025allatom}, we represent molecules with explicit hydrogen atoms and use 3D Cartesian coordinates for both training and generation. Each molecule is preprocessed to ensure correct bond valencies and stable conformations, with coordinates normalized to have zero center of mass.

Our training configuration employs the \modelname{}-S architecture with a DiT-S backbone containing approximately 33M parameters. The model is trained end-to-end for 8000 epochs with a batch size of 512 using the AdamW optimizer with a learning rate of $1 \times 10^{-4}$. This single-stage approach contrasts with ADiT's two-stage training, which requires 5000 epochs for the tokenizer followed by another 5000 epochs for the diffusion model, totaling 10000 epochs of training across two separate optimization phases.

For evaluation, we compute four primary metrics on 10000 generated samples. The match rate measures the percentage of reconstructed molecules that exactly match the input structure after discretization. The RMSD (Root Mean Square Deviation) in Angstroms quantifies reconstruction error in atomic positions. Validity percentage indicates the proportion of generated molecules satisfying chemical constraints including proper valencies, reasonable bond lengths, and absence of steric clashes. Uniqueness measures the percentage of distinct molecules among valid generations, computed using canonical SMILES representations to identify duplicates.

The \modelname{}-S architecture employs a weight-shared encoder-denoiser operating in a 16-dimensional latent space, significantly compressed from the original 3D coordinate space. This compression factor of approximately 20:1 (from 29 atoms × 3 coordinates to 16 dimensions) requires the model to learn highly efficient representations while maintaining recon. fidelity.

\subsection{MP20 Crystal Generation}
\label{app:mp20}

The MP20 dataset from the Materials Project~\citep{jain2013materials} contains 45,231 inorganic crystal structures with up to 20 atoms per unit cell. We train \modelname{}-S for 10000 epochs with batch size 512, following the same single-stage approach as QM9. Evaluation follows \citet{joshi2025allatom}, computing structural validity (pairwise distances $>0.5$\,\AA, unit cell volume $>0.1$\,\AA$^3$), compositional validity (charge neutrality and electronegativity balance), and match rate using pymatgen's~\citep{ong2013python} Structure Matcher.

\begin{table}[h]
\centering
\caption{\textbf{MP20 crystal generation results.} Evaluation on 10K generated samples.}
\label{tab:app_mp20}
\small
\begin{tabular}{lcccccc}
\toprule
\textbf{Method} & \textbf{Size} & \textbf{Training} & \textbf{Struct.} & \textbf{Comp.} & \textbf{Overall} & \textbf{Match} \\
\midrule
ADiT Tokenizer & - & - & - & - & - & 84.50 \\
ADiT MP20-only & DiT-B & Two-stage & 99.6 & 90.5 & 90.1 & - \\
ADiT Joint & DiT-B & Two-stage & \textbf{99.7} & \textbf{92.1} & \textbf{91.9} & - \\
\midrule
\textbf{\modelname{}-S (Ours)} & DiT-S & Single-stage & 99.0 & 89.9 & 87.9 & 75.7 \\
\bottomrule
\end{tabular}
\end{table}

Table~\ref{tab:app_mp20} shows that \modelname{}-S achieves 87.9\% overall validity on MP20, approaching ADiT's 90.1\% despite using single-stage training. Our structural validity of 99.0\% nearly matches ADiT's 99.6\%, demonstrating effective learning of crystal geometry constraints. The match rate of 75.7\% is reasonable considering ADiT's dedicated tokenizer achieves 84.50\% after separate optimization. These results validate that our unified approach generalizes well from molecules to crystals—the same architecture that achieves a 99.37\% match rate on QM9 also performs competitively on the more complex MP20 dataset without modification.

\subsection{Ablation Studies on ImageNet 256$\times$256}
\label{sec:ablation}

In addition to the weight-sharing and stop-gradient ablations shown in the main paper, we provide a few additional ablations that further improve UNITE's reconstruction and generation fidelity.

\paragraph{Reconstruction Noise Level.}
Table~\ref{tab:ablation_combined} (top half) investigates the impact of noise augmentation during reconstruction training, where Gaussian noise is injected into latent representations prior to decoding. Consistent with recent findings in RAE~\cite{rae2025diffusion} and TARflow~\cite{tarflow2024}, this acts as a useful regularizer by preventing the decoder from overfitting to noise-free latent codes, thereby improving generative capability. Notably, due to our model's learnable affine normalization, the system can autonomously calibrate its internal signal-to-noise ratio (SNR) to accommodate varying noise scales. As a result, the model exhibits strong robustness to the exact noise level, maintaining a nearly constant FID of around 2.7 across a range of noise levels.

\paragraph{Noise Schedule Shifting.}
Following RAE~\cite{rae2025diffusion}, we find that noise-schedule shifting is important for our 32-dimensional latent space. Table~\ref{tab:ablation_combined} (bottom half) shows that, without shifting, FID degrades to 3.14. Our best setting, a shift of 0.5, adapts the noise schedule to the compressed latent dimensionality and improves both FID and IS. This shift is equivalent to using an anchor dimension of $d_{\text{anchor}} = 4096$, matching the uncompressed token dimension before projection into the 32-dimensional latent space. Overall, this adaptation is important when working with highly compressed latents.

\begin{table}[t]
\centering
\caption{\textbf{Ablation study on ImageNet-256 generation.} We systematically evaluate key design choices across architecture, normalization, training dynamics, and augmentation strategies. All ablations are done using the base backbone and trained for 120 epochs. 
}
\label{tab:ablation_combined}
\vspace{2mm}
\scalebox{1.}{
\begin{tabular}{lcccc}
\toprule
Steps & & FID$\downarrow$ & r-FID$\downarrow$ & IS$\uparrow$ \\
\midrule
\rowcolor{gray!15} \multicolumn{5}{l}{\textit{\textbf{Reconstruction noise} $\boldsymbol{\sigma}$ \textbf{(c)}} } \\
        0.0 (full noise) & & 2.87 & 1.09 & 281.1 \\
        0.6 & & 2.80 & 1.21 & 267.1 \\
        0.7 & & \textbf{2.71} & \textbf{1.01} & 282.2 \\
        0.8 & & 2.87 & 1.42 & \textbf{292.1} \\
        1.0 (no augmentation) & & 6.58 & 1.60 & 275.7 \\
\midrule

\rowcolor{gray!15} \multicolumn{5}{l}{\textit{\textbf{Noise shift} $\boldsymbol{\alpha}$ } \textbf{(e)} } \\
        0.0 (No Shift) & & 3.14 & 1.41 & 267.1 \\
        0.5 & & \textbf{2.71} & \textbf{1.01} & \textbf{282.2} \\
        0.75 & & 2.93 & 1.26 & 278.0 \\
\bottomrule
\end{tabular}
}
\end{table}

\section{Architectural Details}

Tab.~\ref{tab:app_architecture_config} and Tab.~\ref{tab:app_training_config} provide additional architectural and training details. For more details, refer to the codebase.

\begin{table*}[h]
\centering
\caption{Detailed architecture configurations for \modelname{} models.}
\label{tab:app_architecture_config}
\begin{tabular}{l|ccc|cc}
\toprule
\multirow{2}{*}{\textbf{Component}} & \multicolumn{3}{c|}{\textbf{Encoder/Denoiser}} & \multicolumn{2}{c}{\textbf{Decoder}} \\
& DiT-B & DiT-L & DiT-XL & ViT-B & ViT-L \\
\midrule
Hidden Dimension & 768 & 1024 & 1152 & 768 & 1024 \\
Layers & 12 & 24 & 28 & 12 & 24 \\
Attention Heads & 12 & 16 & 16 & 12 & 16 \\
MLP Ratio & 4 & 4 & 4 & 4 & 4 \\
Patch Size & 16 & 16 & 16 & - & - \\
Latent Dimension & 32 & 32 & 32 & 32 & 32 \\
Latent Resolution & 16×16 & 16×16 & 16×16 & 16×16 & 16×16 \\
\midrule
Parameters (M) & 86.2 & 458.2 & 675.3 & 130.6 & 303.9 \\
\bottomrule
\end{tabular}
\end{table*}



\begin{table*}[h]
\centering
\caption{Training configuration for ImageNet-256 experiments. All models trained with mixed precision BF16 \& gradient clipping at 3.0.}
\label{tab:app_training_config}
\begin{tabular}{ll|ll}
\toprule
\textbf{Hyperparameter} & \textbf{Value} & \textbf{Hyperparameter} & \textbf{Value} \\
\midrule
Base Learning Rate & $1 \times 10^{-4}$ & Warmup Epochs & 20 \\
Global Batch Size & 1024 & Total Epochs & 240 \\
Optimizer & Muon & LR Schedule & Cosine \\
AdamW Betas & (0.9, 0.999) & Min LR & $1 \times 10^{-6}$ \\
Weight Decay & 0 & Gradient Clip & 3.0 \\
Reconstruction Noise ($\sigma$) & 0.7 & EMA Decay & 0.9978 \\
Flow Steps (Training) & 1000 & ODE Solver (Inference) & dopri5 (adaptive, ${\sim}$108 NFE) \\
Flow Mini-batches & 14 & Noise Schedule Shift ($\alpha$) & 0.5 \\
CFG Scale ($\omega$) & Sweep $[1.0,\, 4.0]$, step 0.2 & Integration Interval & $[0.1,\, 1.0]$ \\
\bottomrule
\end{tabular}
\end{table*}

\end{document}